# A Unifying Contrast Maximization Framework for Event Cameras, with Applications to Motion, Depth, and Optical Flow Estimation

Guillermo Gallego, Henri Rebecq, Davide Scaramuzza
Dept. of Informatics and Neuroinformatics, University of Zurich and ETH Zurich

## Abstract

*We present a unifying framework to solve several computer vision problems with event cameras: motion, depth and optical flow estimation. The main idea of our framework is to find the point trajectories on the image plane that are best aligned with the event data by maximizing an objective function: the contrast of an image of warped events. Our method implicitly handles data association between the events, and therefore, does not rely on additional appearance information about the scene. In addition to accurately recovering the motion parameters of the problem, our framework produces motion-corrected edge-like images with high dynamic range that can be used for further scene analysis. The proposed method is not only simple, but more importantly, it is, to the best of our knowledge, the first method that can be successfully applied to such a diverse set of important vision tasks with event cameras.*

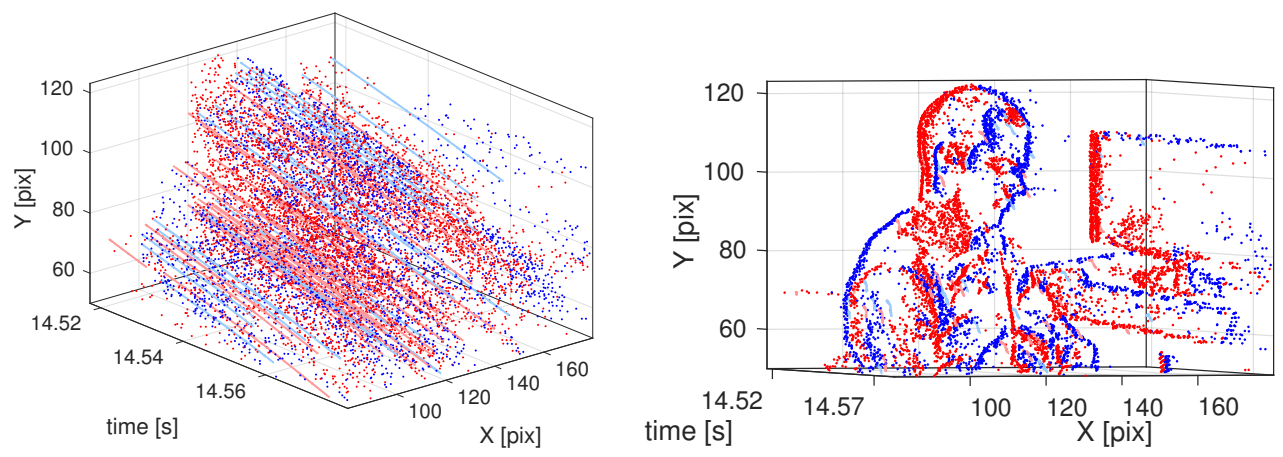

(a) (b)

Figure 1: (a) Events (dots) caused by a moving edge pattern and point trajectories in a space-time region of the image plane, colored according to event polarity (blue: positive event, i.e., brightness increase; red: negative event, i.e., brightness decrease). (b) Visualization of the events along the direction of the *point trajectories* highlighted in (a); corresponding events line up, revealing the edge pattern that caused them. Our approach works by maximizing the contrast of an image of warped events similar to (b). A video demonstrating our framework is available at: https://youtu.be/KFMZFhi-9Aw

## 1. Introduction

Unlike traditional cameras, which produce intensity images at a fixed rate, event cameras, such as the Dynamic Vision Sensor (DVS) [1], have independent pixels that report only intensity changes (called "events") asynchronously, at the time they occur. Each event consists of the spatio-temporal coordinates of the brightness change (with microsecond resolution) and its sign[1]. Event cameras have several advantages over traditional cameras: a latency in the order of microseconds, a very high dynamic range (140 dB compared to 60 dB of traditional cameras), and very low power consumption (10 mW vs 1.5 W of traditional cameras). Moreover, since all pixels capture light independently, such sensors do not suffer from motion blur. In summary, event cameras represent a paradigm shift since visual information is: (*i*) sampled based on the dynamics of the scene, not based on an external clock[2], and (*ii*) encoded using an asynchronous and sparse stream of events, which is fundamentally different from a sequence of images. Such a paradigm shift calls for new methods to process visual information and unlock the capabilities of these novel sensors.

Algorithms for event cameras can be classified according to different criteria. Depending on the way in which events are processed, two broad categories can be distinguished: *1)* methods that operate on an *event-by-event basis*, where the state of the system (the estimated unknowns) can change upon the *arrival of a single event* [2, 3, 4, 5, 6, 7, 8, 9, 10, 11, 12, 13, 14, 15]; and *2)* methods that operate on *groups of events*. This category can be further subdivided into two: *2a)* methods that *discard temporal information* of the events and accumulate them into frames to re-utilize traditional, image-based computer vision algorithms [16, 17, 18], and *2b)* methods that *exploit the temporal information* of the events during the estimation process, and therefore cannot re-utilize traditional computer vision algorithms (sample applications of these methods include variational optical flow estimation [19], event-based multi-view stereo [20, 21], rotational motion estimation [22], feature tracking [23], pose estimation [24] and visual-inertial odometry [25, 26, 27]).

[1] An animation of the principle of operation of event cameras can be found in the video of [2] https://youtu.be/LauQ6LWTkxM?t=25

[2] If nothing moves in the scene, no events are generated. Conversely, the number of events (samples) increases with the amount of scene motion.

Event-by-event–based methods rely on the availability of additional appearance data, in the form of grayscale images or a photometric map of the scene, which may be built from past events or provided by additional sensors. Then, each incoming event is compared against such appearance data and the resulting mismatch is used to update the system unknowns. In contrast, methods that operate on groups of events do not rely on prior appearance data. Instead, they aggregate the information contained in the events to estimate the unknowns of the problem. Since each event carries little information and is subject to noise, several events must be processed together to yield a sufficient signal-to-noise ratio for the problem considered.

Both categories present methods with advantages and disadvantages and current research focuses on exploring the possibilities that each category has to offer. Filters, such as the Extended Kalman filter, are the dominant framework of the event-by-event–based type of methods. In contrast, for the groups-of-events–based category, only *ad-hoc solutions* for every problem have been proposed. We present the first unifying framework for processing groups of events while exploiting their temporal information (i.e., category *2b*).

**Contribution.** This paper presents the first unifying framework that allows to tackle several important estimation problems for event cameras in computer vision. In a nutshell, our framework seeks for the point trajectories on the image plane that best fit the event data, and, by doing so, is able to recover the parameters that describe the relative motion between the camera and the scene. The method operates on groups of events, exploiting both their spatio-temporal and polarity information to produce accurate results. In contrast to event-by-event–based approaches, our method does not rely on additional appearance information and it can be used both for estimation problems with very short characteristic time (optical flow) as well as for problems with longer estimation time (monocular depth estimation). Moreover, our framework implicitly handles data association between events, which is a central problem of event-based vision. Additionally, the framework produces motion-corrected event images, which approximate the image gradients that caused the events. These images can serve as input to more complex processing algorithms such as visual-inertial data fusion, object recognition, etc.

The rest of the paper is organized as follows. Section 2 illustrates the main idea behind our framework on a simple example: optical flow estimation. Then, we generalize it and apply it to other problems, such as depth estimation (Section 3.1), rotational motion estimation (Section 3.2) and motion estimation in planar scenes (Section 3.3). Section 4 concludes the paper.

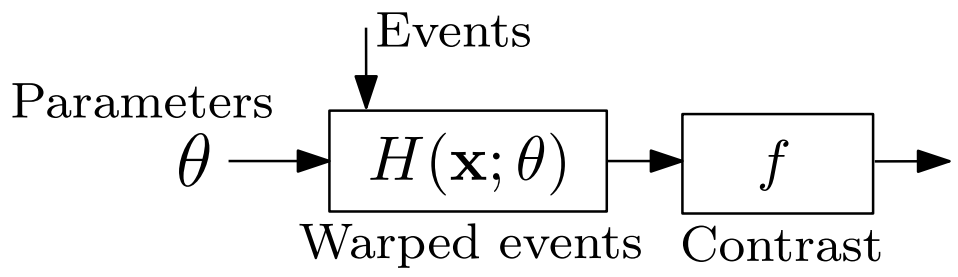

Figure 2: Events are warped according to point trajectories described by motion parameters $\theta$, resulting in an image of warped events $H(\mathbf{x};\theta)$. The contrast of $H$ measures how well events agree with the candidate point trajectories.

## 2. Contrast Maximization Framework

Since event cameras report only pixel-wise brightness changes, it implies that, assuming constant illumination, there must exist (*i*) relative *motion* between the camera and the objects in the scene and (*ii*) sufficient texture (i.e., brightness gradients) for events to be generated. Hence, event cameras respond to the apparent motion of edges. In the absence of additional information about the appearance of the scene that caused the events, the problem of extracting information from the events becomes that of establishing correspondences between them, also known as data association, i.e., establishing which events were triggered by the same scene edge. Since moving edges describe point trajectories on the image plane, we expect corresponding events to be triggered along these trajectories. Fig. 1 illustrates this idea with a simple example where the point trajectories are nearly straight lines. We propose to find the point trajectories that best fit the event data, as in Fig. 1b. Let us describe our framework using a simple yet important example (optical flow estimation) and then let us generalize it to other estimation problems.

### 2.1. Example: Optical Flow Estimation

Assume that we are given a set of events $\mathcal{E} \doteq \{e_k\}_{k=1}^{N_e}$ in a spatio-temporal neighborhood of a pixel, as in Fig. 1a, and the goal is to estimate the optical flow (i.e., motion vector) at that pixel based on the information contained in the events. Recall that each event $e_k \doteq (x_k, y_k, t_k, p_k)$ consists of the space-time coordinates of a predefined brightness change together with its polarity $p_k \in \{-1, +1\}$ (i.e., the sign of the brightness change).

As is standard, optical flow is measured over a small time interval (ideally infinitesimal), and the trajectories followed by points on the image plane are locally straight, approximated by translations: $\mathbf{x}(t) = \mathbf{x}(0) + \mathbf{v}t$, where $\mathbf{x} \doteq (x,y)^\top$ and $\mathbf{v}$ is the velocity of the point (i.e., the optical flow). Hence, we expect corresponding events (triggered by the same edge) to lie on such trajectories (Fig. 1b).

Our framework, summarized in Fig. 2, consists of counting the events or summing their polarities along the straight trajectories given by a candidate optic flow and computing the variance ($f$) of the resulting sums ($H$), which measures how well the events agree with the candidate trajectories.

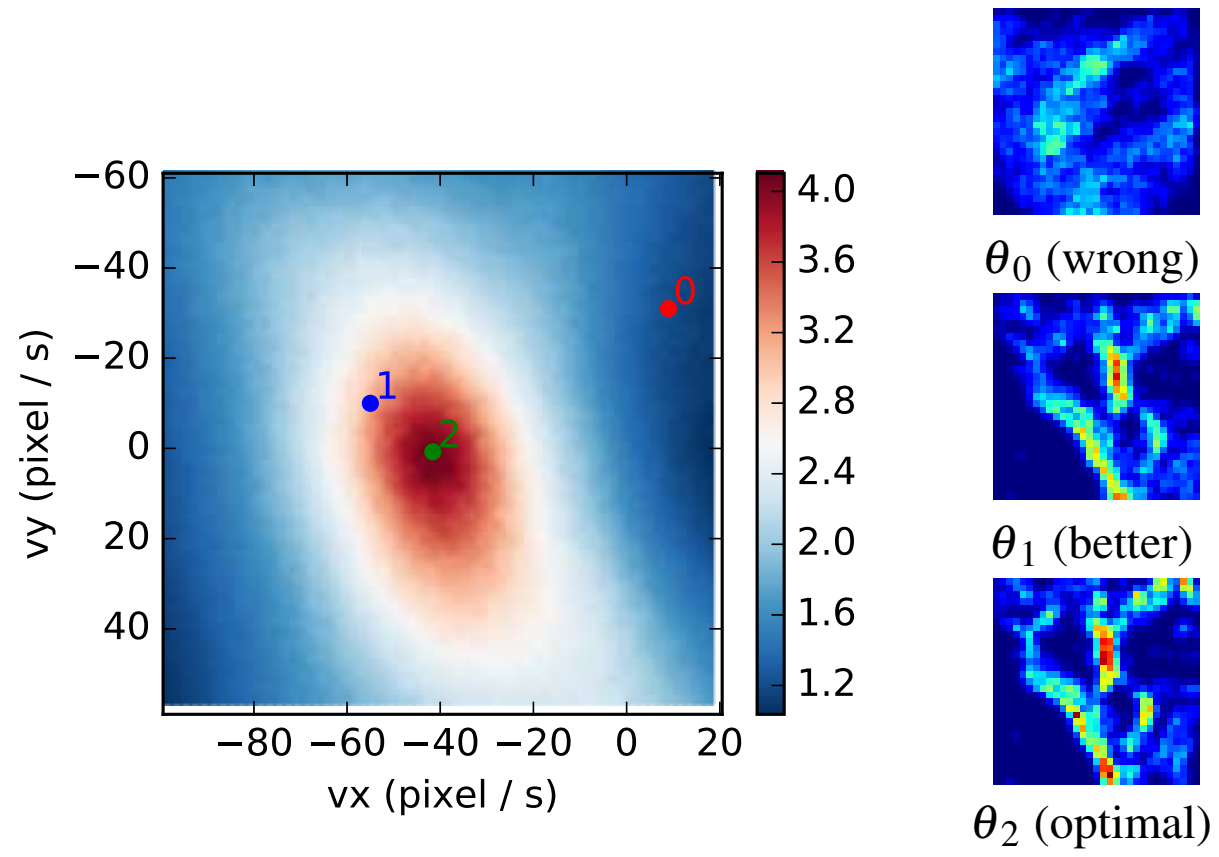

(a) $f(\theta)$ as a function of the optical flow $\theta \equiv \mathbf{v}$. For illustration, we show in (b) the associated images of warped events $H$ for three candidate velocities.

(b) Warped events $H(\mathbf{x};\theta)$, for $\theta_i$, $i = 0,1,2$ in (a).

Figure 3: *Optical Flow* (patch-based) estimation. The correct optical flow can be found as the one that maximizes the contrast (Fig. 3a) of the images of warped events (Fig. 3b).

More specifically, to sum the events, we warp them to a reference time $t_{\text{ref}}$ using the proposed trajectories. Following the local-flow–constancy hypothesis [28], we assume that the flow is constant in the space-time neighborhood spanned by the events, and warp the events, $e_k \mapsto e'_k$, according to

$$\mathbf{x}'_k \doteq \mathbf{W}(\mathbf{x}_k, t_k; \theta) = \mathbf{x}_k - (t_k - t_{\text{ref}})\theta, \qquad (1)$$

with $\theta = \mathbf{v}$ the candidate velocity. Then, we build an image patch of warped events:

$$H(\mathbf{x};\theta) \doteq \sum_{k=1}^{N_e} b_k\, \delta(\mathbf{x} - \mathbf{x}'_k), \qquad (2)$$

where each pixel $\mathbf{x}$ sums the values $b_k$ of the warped events $\mathbf{x}'_k$ that fall within it (indicated by the Dirac delta $\delta$). If $b_k = p_k$, the polarities of the events along the trajectories are summed; whereas, if $b_k = 1$, the number of events along the trajectories are computed. In the rest of the paper, we use $b_k = 1$ and show comparisons with $b_k = p_k$ in the supplementary material. Finally, we compute the variance of $H$, which is a function of $\theta$,

$$f(\theta) = \sigma^2\big(H(\mathbf{x};\theta)\big) \doteq \frac{1}{N_p}\sum_{i,j}(h_{ij} - \mu_H)^2, \qquad (3)$$

where $N_p$ is the number of pixels of $H = (h_{ij})$ and $\mu_H \doteq \frac{1}{N_p}\sum_{i,j} h_{ij}$ is the mean of $H$.

Fig. 3a shows the variance (3) as a heat map, for a given set of events $\mathcal{E}$. Observe that it is smooth and has a clear peak. The images of warped events corresponding to three different motion vectors $\theta_i$ in Fig. 3a are displayed in Fig. 3b (represented in pseudo-color, from blue (few events) to red (many events)). As can be seen, the warped events are best aligned in the image $H$ that exhibits highest variance, i.e., highest contrast (and sharpness), attained at $\theta^* = \arg\max_\theta f(\theta) \approx (-40, 0)^\top$ pixel/s. Hence, our strategy for optical flow estimation consists of seeking the parameter $\theta$ that maximizes (3).

As anticipated in Section 1, our method produces motion-corrected image patches $H$, which approximate the gradients of the brightness pattern that caused the events (also illustrated in Fig. 1b). More specifically, $H$ represents the brightness increment along the candidate trajectories. For optimal trajectories, this increment is proportional to $\nabla I \cdot \mathbf{v}$, the dot product of the brightness gradient and the motion vector (due to the optical flow constraint). These motion-corrected image patches can be useful for feature tracking, such as [27, 29].

Our framework implicitly defines data association between the events. Replacing the delta in (2) with a smooth approximation, $\delta(\mathbf{x}) \approx \delta_\varepsilon(\mathbf{x})$, such as a Gaussian, $\delta_\varepsilon(\mathbf{x} - \mu) \doteq \mathcal{N}(\mathbf{x}; \mu, \varepsilon^2 \texttt{Id})$ (we use $\varepsilon = 1$ pixel), we see that every warped event $e'_k$ influences every other event $e'_n$, and the amount of influence is given by $\delta_\varepsilon(\mathbf{x}'_n - \mathbf{x}'_k)$, which, in the case of a Gaussian, is related to the Euclidean distance $\|\mathbf{x}'_n - \mathbf{x}'_k\|$. Hence, our method has a built-in soft data association between all events, implicitly given by a function of the distance between them: the further away warped events are, the less likely they are corresponding events.

## 2.2. General Description of the Framework

The example in the previous section contains all the ingredients of our event-processing framework. Let us now describe it in a more generic manner, to apply to other estimation problems (in Section 3).

We propose to find the point trajectories on the image plane that best fit the event data. More specifically, assume we are given a set of events, $\mathcal{E} \doteq \{e_k\}_{k=1}^{N_e}$, typically continuous in time, as in Fig. 1, acquired while the camera and/or the scene undergo some motion for which we have a geometric model of how points move on the image plane. Such a geometric model depends on the particular estimation problem addressed (optical flow, depth estimation, motion estimation). The goal is to estimate the parameters of the model based on the information contained in the events. We assume that estimation is possible, in that the model parameters (unknowns) are shared among multiple events (fewer parameters than events) and are observable. To solve the problem, we build candidate point trajectories $\mathbf{x}(t)$ according to the motion and scene models, and measure the goodness of fit between these trajectories and the event data $\mathcal{E}$ using an objective function (3) (see Fig. 2). Then, we use an optimization algorithm to seek for the point trajectories (i.e., the parameters $\theta$ of the motion and scene models) that maximize the objective function. As shown in Fig. 3b, good trajectories are those that align corresponding events, and so the objective function that we propose measures how well events are aligned along the candidate trajectories. There are two by-products of our framework: (*i*) the estimated point trajectories implicitly establish correspondences between events (i.e., data association), (*ii*) the trajectories can be used to correct for the motion of the edges.

**2.2.1. Steps of the Method.** Our method consists of three main steps:

1. Warp the events into an image $H$, according to the point trajectories defined by the above-mentioned geometric model and candidate parameters $\boldsymbol{\theta}$.
2. Compute a score $f$ based on the image of warped events.
3. Optimize the score or objective function with respect to the parameters of the model.

In step 1, events are geometrically transformed taking into account their space-time coordinates and other known quantities of the point-trajectory model, $e_k \mapsto e'_k(e_k;\boldsymbol{\theta})$, resulting in a set of warped events $\mathcal{E}' \doteq \{e'_k\}_{k=1}^{N_e}$. The warp, such as $\mathbf{W}$ in (1), transports each event along the point trajectory that passes through it, until a reference time is reached (e.g., the time of the first event): $e_k \doteq (x_k, y_k, t_k, p_k) \mapsto (x'_k, y'_k, t_{\text{ref}}, p_k) \doteq e'_k$.

In step 2, an image or histogram of warped events $H(\mathcal{E}')$ is created (using their polarities $p_k$ or their count), and an objective function (a measure of dispersion) is computed, $f(H(\mathcal{E}'))$. We use as dispersion metric the variance of $H$, which is known as *contrast* in image processing terminology, and we seek to maximize it. The objective function represents the statistics of the warped events $\mathcal{E}'$ according to the candidate model parameters $\boldsymbol{\theta}$, hence it measures the goodness of fit of $\boldsymbol{\theta}$ to the event data $\mathcal{E}$.

In step 3, an optimization algorithm, such as gradient ascent or Newton's method, is applied to obtain the best model parameters, i.e., the point trajectories on the image plane that best explain the event data. The framework is flexible, not relying on any specific optimizer.

**2.2.2. Contrast Maximization.** By maximizing the variance of the image of warped events $H(\mathcal{E}'(\boldsymbol{\theta}))$ we favor the point trajectories that accumulate (i.e., align) the warped events on the image plane. The accumulation of warped events in some regions and the dispersion of events in other regions (since the total number of events $N$ is constant) produces an image $H$ with a larger range, and, therefore a higher contrast, which is clearly noticeable if $H$ is displayed in grayscale, and hence the name *contrast maximization* framework. In essence, the goal of the optimization framework is to "pull apart" the statistics (e.g., polarity) of the regions with and without events, in a similar way to the segmentation approach in [30].

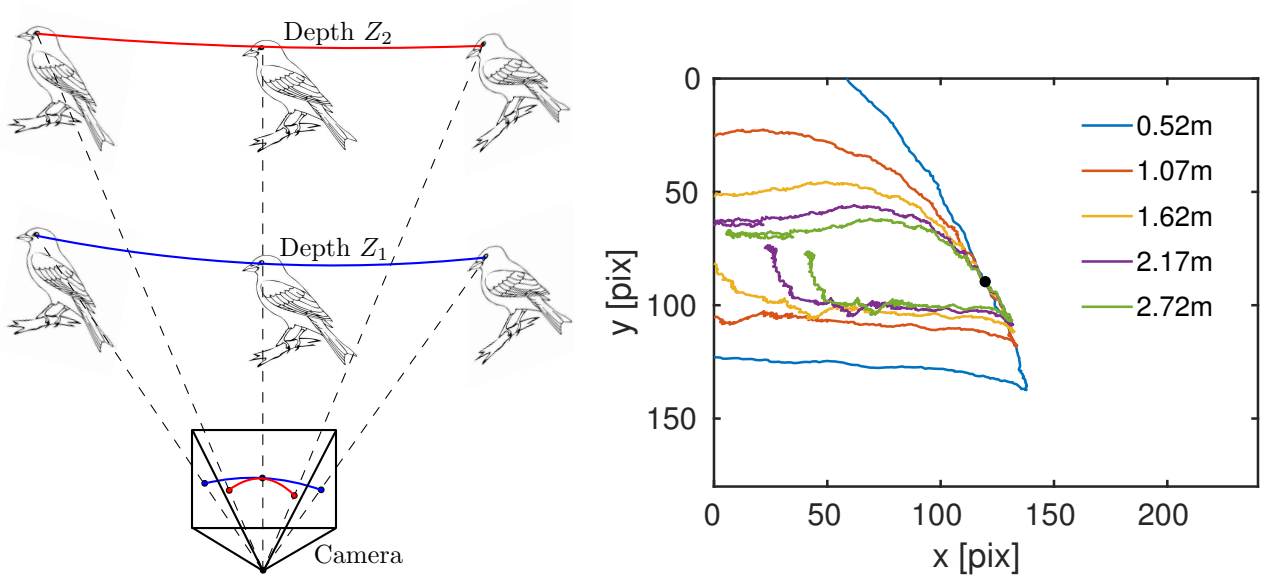

Figure 4: *Depth Estimation*. Left: trajectories of a 3D point (the eye of the bird) relative to the camera. The point closer to the camera has larger apparent motion, and therefore typically describes longer trajectories on the image plane. Right: Trajectories of an image point (the image center, in black), for different depth values with respect to a reference view while the camera undergoes a 6-DOF motion. Each depth value produces a different point trajectory; yielding a 1D family of curves. This is an example of a 2 s segment from a sequence of the Event Camera Dataset [31].

Contrast is related to image sharpness, and therefore, an observed effect is that the images of warped events with higher contrast are also sharper (see Fig. 3b), which is consistent with the better alignment of the warped events. The image of warped events associated to the optimal parameters is a motion-corrected edge-like image, where the "blur" (trace of events) due to the moving edges has been removed (cf. Fig. 3b top and bottom). Such edge-like image represents the brightness-increment patterns causing the events.

**2.2.3. Computational Complexity.** The core of our method is the computation of the image of warped events (2), whose computational complexity is linear on the number of events to be warped. The computation of the contrast (3) is typically negligible compared to the effort required by the warp. The overall cost of the method also depends on the choice of algorithm used to maximize the contrast, which is application-dependent.

# 3. Sample Applications

Our framework is flexible and generic. In this section, we use it to solve various important problems in vision.

## 3.1. Depth Estimation

Consider the scenario of an event camera moving in a static scene and the goal is to infer depth. That is, consider the problem of event-based multi-view stereo (EMVS) [21, 32] (3D reconstruction) from a set of events $\mathcal{E} = \{e_k\}_{k=1}^{N_e}$. By assumption of the problem, the pose of the event camera $\mathtt{P}(t)$ is known for every time $t$, where $\mathtt{P}$ denotes the projection matrix of the camera. We assume that the intrinsic parameters of the camera are also known[3] and that lens distortion has been removed. Following the framework in the previous section, we first specify the geometry of the problem, the warp (i.e., point trajectories) and the score function.

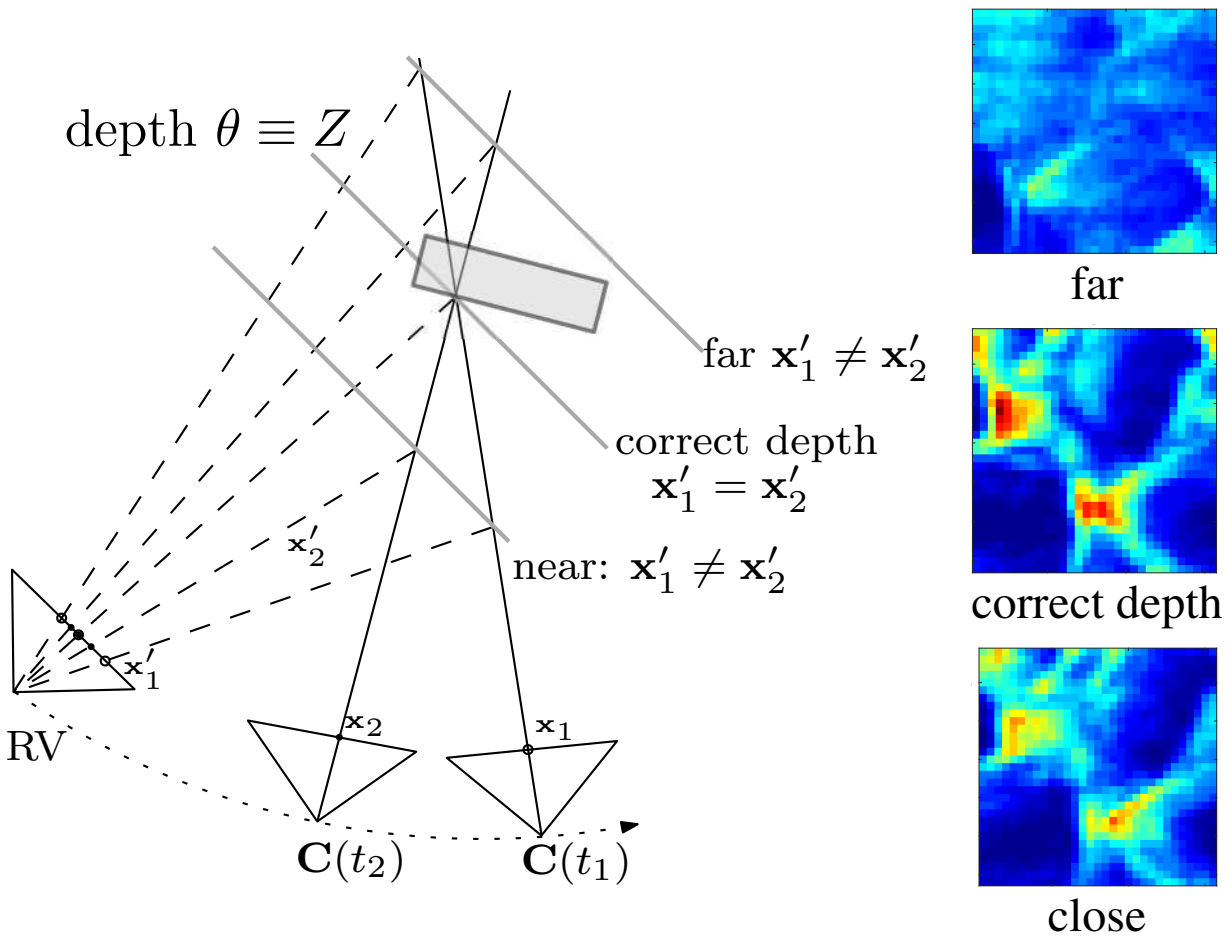

(a) Geometric configuration of the problem (b) Warped events

Figure 5: *Depth Estimation*. Alignment of warped events $\mathbf{x}'_k(\theta)$, for different depth values $\theta \equiv Z$, measured with respect to the reference view (RV). In (b), the patch of warped events (2) is represented in pseudo-color, from few events aligned (blue) to many events (red). At the correct depth, the patch has the highest variance (i.e., image contrast).

In this scenario, the trajectory of an image point obtained by projection of a 3D point is parametrized by the known 6-DOF motion of the camera and the depth of the 3D point with respect to a reference view. This yields a 1D family of curves in the image plane, parametrized by depth. Each depth value gives a different curve followed by the point in the image plane. This is illustrated in Fig. 4. As in the Space-Sweep approach of [21], we consider a reference view provided by a virtual camera at some location, for example, at a point along the trajectory of the event camera (see Fig. 5). Let us formulate the problem for a small patch in the reference view. For simplicity, we assume that all the points within the patch have the same depth, which is some candidate value $\theta = Z$. The three main steps of our method (Section 2.2.1) are as follow:

1. (a) Transfer the events (triggered at the image plane of the moving event camera) onto the reference view using the candidate depth parameter, as illustrated in Fig. 5. An event $e_k$ is transferred, using the warp ($\mathbf{W}$), onto the event $e'_k = (\mathbf{x}'_k, t_{\text{ref}}, p_k)$ with

$$\mathbf{x}'_k = \mathbf{W}(\mathbf{x}_k, \mathtt{P}(t_k), \mathtt{P}_v; \theta), \tag{4}$$

where $\mathtt{P}(t)$ is the pose of the event camera at time $t$ and $\mathtt{P}_v \doteq \mathtt{P}(t_{\text{ref}})$ is the pose of the virtual camera. The warp is the same as in space-sweep multi-view stereo: points are transferred using the planar homography [35, ch. 13] induced by a plane parallel to the image plane of the reference view and at the given depth (see Fig. 5).

   (b) Create an image (patch) of warped events (2) by counting the number of events along the candidate point trajectories (e.g., Fig. 4).
2. Measure the goodness of fit between the events and the depth value $\theta$ by means of the variance (i.e., contrast) of the image of warped events (3).
3. Maximize the contrast by varying the depth parameter $\theta$.

Fig. 5 illustrates the above steps. In Fig 5a an event camera with optical center $\mathbf{C}(t)$ moves in front of a scene with an object (gray box). Two events $e_i = (\mathbf{x}_i, t_i, p_i), i = \{1,2\}$ are transferred from the event camera to a reference view (RV) via the warp (4) using three candidate depth values (in front of, at the object and behind it, respectively). The points transferred using depth values in front and behind the object are not aligned, $\mathbf{x}_1 \neq \mathbf{x}_2$, whereas the points transferred using the correct value are aligned, $\mathbf{x}_1 = \mathbf{x}_2$. Event alignment is more noticeable in Fig. 5b, with patches of warped events (2) from a sequence of the Event Camera Dataset [31]. At the correct depth, the warped events present a better alignment, and therefore higher contrast (and sharpness) of $H$ (here represented in pseudo-color instead of grayscale), compared to the cases of wrong depth values (labeled as "close" and "far").

Fig. 6 shows depth estimation for two patches in a sequence from the dataset [31]. The sequence was recorded with a DAVIS camera [33], which outputs asynchronous events and grayscale frames (at 24 Hz). The frame in Fig. 6a is only shown for visualization purposes; our method processes solely the events. Fig. 6b shows how the contrast of the warped events (3) varies with respect to the depth parameter $\theta$, for each patch in Fig. 6a. The actual warped events (2) at selected depth values are displayed in Fig. 6c. Remarkably, the contrast curves (Fig. 6b) have a smooth variation, with a clear maximum at the correct depth value. The warped events in (2) show, indeed, a better alignment (i.e., higher contrast) at the correct depth than at other depth values. As in Fig. 5b, a pseudo-color scale is used to represent the pixels of $H$, from few events aligned (blue) to many events (red). Additionally, note that, although the contrast curves are from different patches, the "spread" of the curves increases with the value of the peak depth, which is consistent with the well-known fact that, in short-baseline stereo, depth uncertainty grows with depth. Finally, observe that

[3] An event camera may be intrinsically calibrated using event-based algorithms, such as [2] or, as in the case of the DAVIS [33], using standard algorithms [34] since the DAVIS comprises both a traditional frame camera and an event sensor (DVS) in the same pixel array.

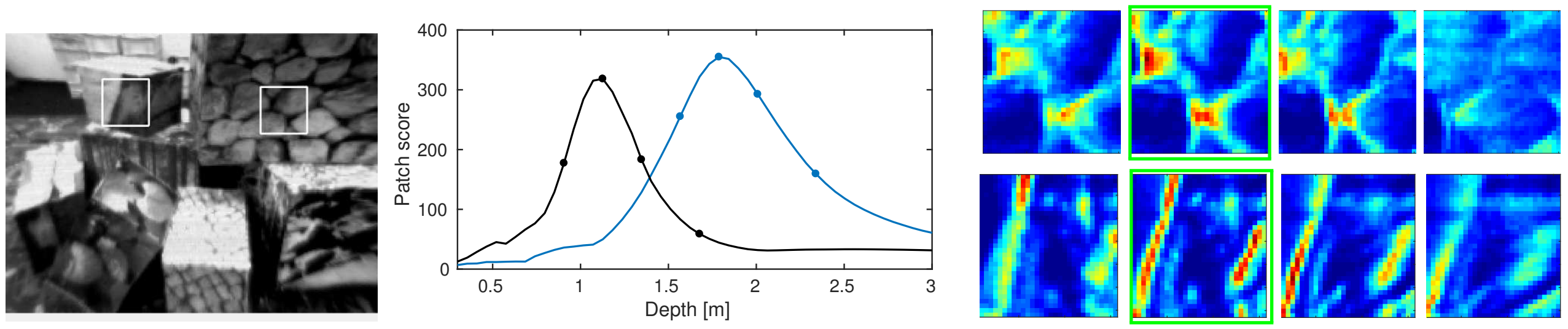

(a) Two patches on the reference view. Monochrome image used only for visualization purposes.

(b) Contrast vs. depth for the events corresponding to the two patches in (a). Contrast consists of the variance (3) of the warped events in (c).

(c) Warped events for the 4 increasing depth values on each curve in (b). Top: right patch in (a), with best depth $\approx 1.1$ m (black line in (b)). Bottom: left patch in (a), with best depth $\approx 1.8$ m (blue line in (b)).

Figure 6: *Depth estimation* by contrast maximization. Two patches are analyzed.

the patches of warped events with highest contrast (highlighted in green in Fig. 6c), resemble the edgemap (gradient magnitude) of the grayscale patches in Fig. 6a, showing that indeed the events are triggered by moving edges and that our framework recovers, as a by-product, an approximation to edgemap of the scene patch that caused the events.

The above procedure yields the depth value at the center of the patch in the reference view. Repeating the procedure for every pixel in the reference camera for which there is sufficient evidence of the presence of an edge produces a semi-dense depth map. This is shown in Fig. 7, using a sequence from the dataset [31]. In this experiment, we computed the contrast (3) on patches of $3 \times 3$ pixels in the reference view; the patches $H$ were previously weighted using a Gaussian kernel, $H(\mathbf{x}) \leftarrow w(\mathbf{x})H(\mathbf{x})$, to emphasize the contribution of the center pixel. A map of the maximum contrast for every pixel of the reference view is shown in Fig. 7b. For visualization purposes, contrast is represented in negative form, from bright (low contrast) to dark (high contrast). This map is used to select the points in the reference view with largest contrast, which are the points for which depth estimation is most reliable. The result is a semi-dense depth map, which is displayed in Fig. 7a, color coded, overlaid on a grayscale frame. To select the points, we used adaptive thresholding [36, p.780] on the contrast map, and we used a median filter to remove spike noise from the depth map. As it can be observed, depth is most reliable at strong brightness edges of the scene. The depth map is visualized as a color-coded point cloud in Figs. 7c and 7d.

If the images of warped events (2) are interpreted as depth slices of the disparity space image (DSI) formed by event back-projection [21], our framework estimates depth by selecting DSI regions with largest local variance (contrast). Moreover, our framework allows to iteratively refine the depth values (e.g., by gradient ascent on the objective function (3)) so that they are continuous, i.e., they are not constrained to be the discrete set of values imposed by a voxelization of the DSI.

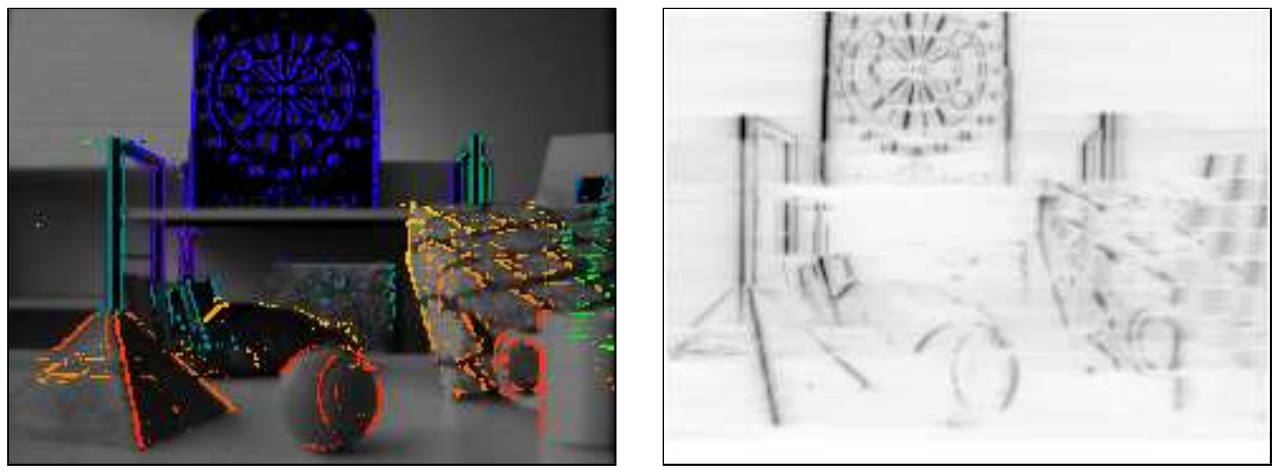

(a) Depth map overlaid on frame.

(b) Contrast map.

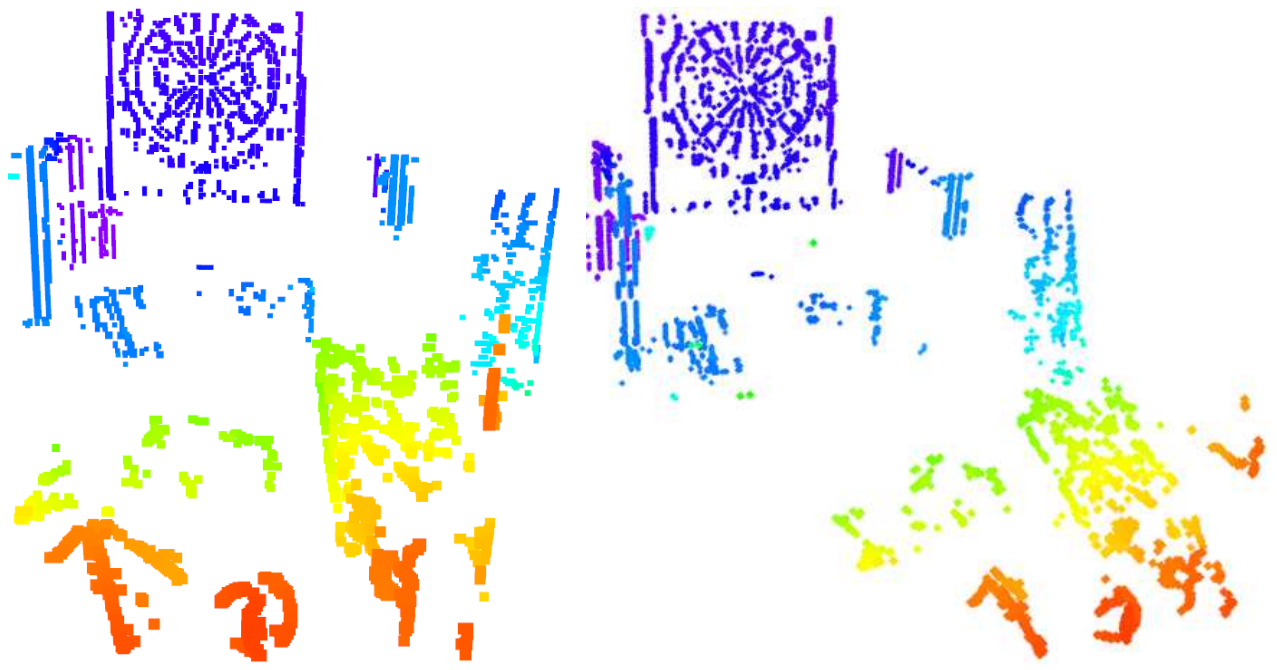

(c) 3D reconstruction, front view

(d) 3D reconstruction, oblique view.

Figure 7: *Depth Estimation*. 3D reconstruction on the `slider_depth` sequence of the dataset [31]. One million events in a timespan of 3.4 s were processed. In (a), (c) and (d), depth from the reference view is color-coded, from red (close) to blue (far), in the range of 0.45 m to 2.4 m. In the supplementary material we show how the reconstruction changes with the number of events processed.

### 3.2. Rotational Motion Estimation

Our framework can also be applied to the problem of rotational motion estimation [9, 22, 37, 38]. Consider the scenario of an event camera rotating in a static scene and the goal is to estimate the camera's ego-motion using the events. As in Section 3.1, assume that the camera is calibrated (known intrinsic parameters and no lens distortion).

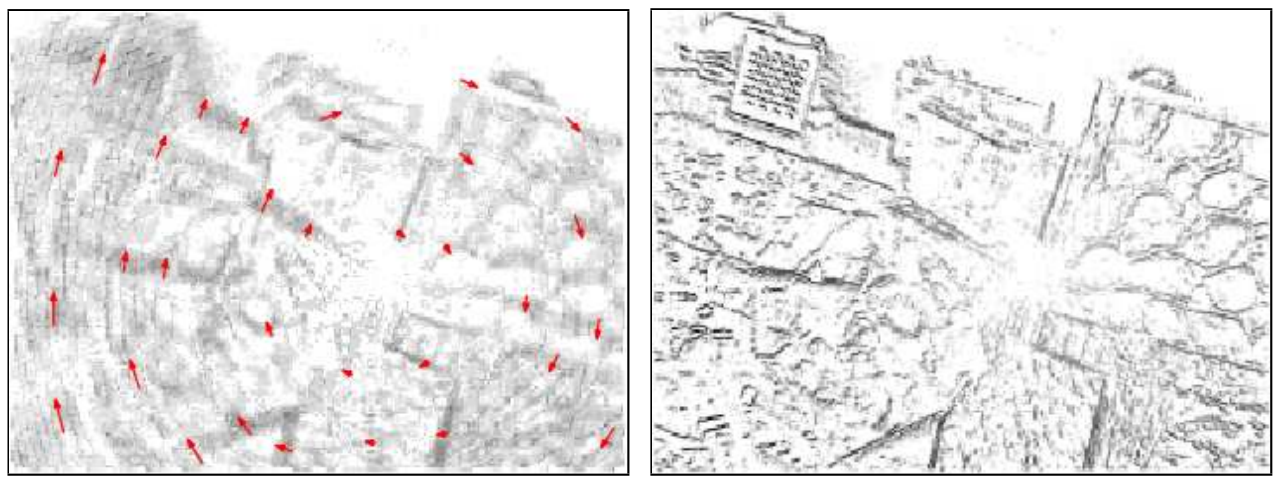

(a) Warped events using $\theta = \mathbf{0}$. Read arrows indicate the true motion of the edges causing the events.

(b) Warped events using estimated angular velocity $\theta^*$, which produces motion-corrected, sharp edges.

Figure 8: *Rotational Motion Estimation*. Images of warped events, displayed in grayscale to better visualize the motion blur due to event misalignment and the sharpness due to event alignment. The direction of the rotation axis is clearly identifiable on the left-image as the point with least motion blur and where fewer events are triggered. Dataset [31].

The type of motion and the motion parameters themselves enforce constraints on the trajectories that image points can follow. For example, in a rotational motion with constant velocity, all point trajectories are parametrized by 3-DOFs: the angular velocity. Our framework aligns events by maximizing contrast over the set of admissible trajectories: those compatible with the rotational motion.

Let us specify the steps of the method (Section 2.2.1) for the problem at hand. Consider all events $\mathcal{E}$ over a small temporal window $[0, \Delta t]$; small enough so that the angular velocity $\boldsymbol{\omega}$ can be considered constant within it, and let $t_{\text{ref}} = 0$. In calibrated coordinates, image points transform according to $\bar{\mathbf{x}}(t) \propto \mathtt{R}(t)\bar{\mathbf{x}}(0)$, where $\bar{\mathbf{x}} \propto (\mathbf{x}^\top, 1)^\top$ are homogeneous coordinates and $\mathtt{R}(t) = \exp(\widehat{\boldsymbol{\omega}} t)$ is the rotation matrix of the (3D) motion [35, p. 204]: exp is the exponential map of the rotation group $SO(3)$ [39] and $\widehat{\boldsymbol{\omega}}$ is the cross-product matrix associated to $\boldsymbol{\omega}$. In step 1, events are warped to $t_{\text{ref}}$ according to the point-trajectory model: $\mathbf{x}'_k = \mathbf{W}(\mathbf{x}_k, t_k; \boldsymbol{\theta})$, with $\boldsymbol{\theta} = \boldsymbol{\omega}$ the angular velocity and

$$\mathbf{W}(\mathbf{x}_k, t_k; \boldsymbol{\theta}) \propto \mathtt{R}^{-1}(t_k)\bar{\mathbf{x}}_k = \exp(-\widehat{\boldsymbol{\theta}} t_k)\bar{\mathbf{x}}_k. \quad (5)$$

The image of warped events is then given by (2). Approaches like [22] use the event polarity, which is indeed beneficial if the motion is monotonic (i.e., does not change direction abruptly). However, as we show, polarity is not needed (see Figs. 8 and 9). In steps 2 and 3, the objective function (3) is maximized using standard optimization algorithms such as non-linear conjugate gradient [40].

Figure 8 shows the result of our method on a group $\mathcal{E}$ of $N_e = 30\,000$ events acquired while the camera is rotating approximately around its optical axis. As it can be seen, our method estimates the motion parameters that remove the motion blur from the image of warped events, providing the sharpest image. Fig. 9 reports the accuracy of our method

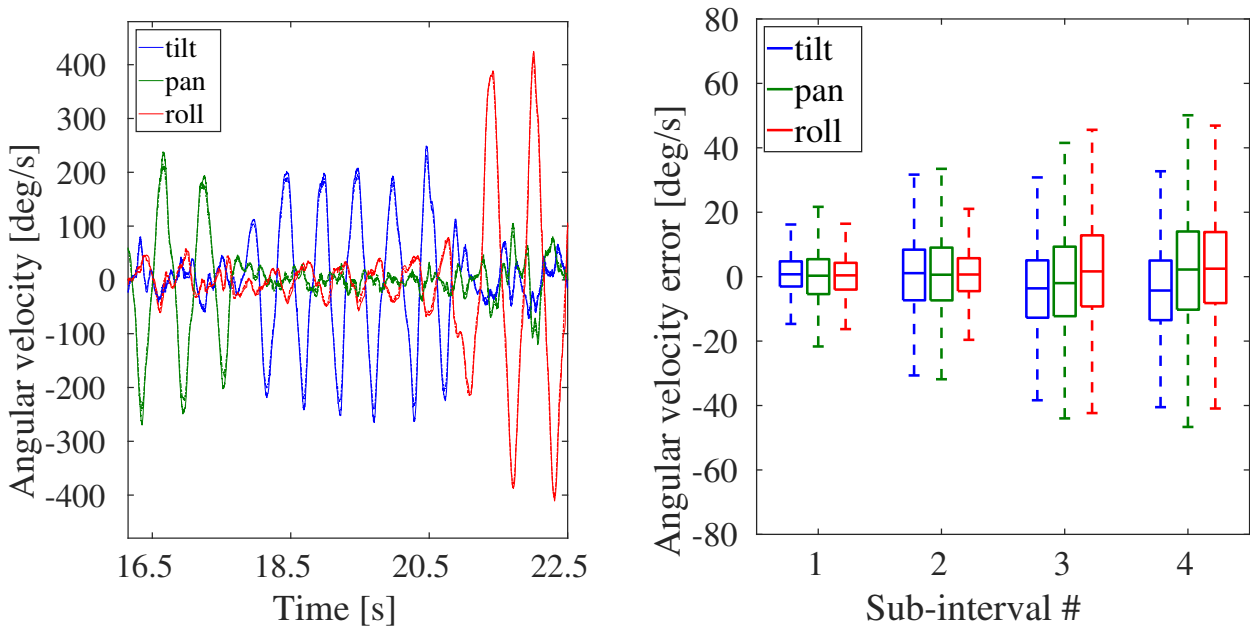

(a) Angular velocity: estimated (solid) vs. ground truth (dashed).

(b) Angular velocity error: estimated vs. ground truth for 4 subintervals of 15 s.

Figure 9: *Rotational Motion Estimation*. Accuracy evaluation. The `boxes_rotation` sequence of the dataset [31] contains 180 million events and reaches speeds of up to 670 °/s. Comparatively, our method produces small errors.

using the same dataset and error metrics as [22]. Both plots compare the recovered rotational motion of the event camera against ground truth. Fig. 9a shows the estimated and ground-truth motion curves, which are almost indistinguishable relative to the magnitude of the motion. Fig. 9b further analyzes the error between them in four subintervals of 15 s with increasing angular velocities (and therefore increasing errors). Our method is remarkably accurate, with RMS errors of approximately 20 °/s with respect to peak excursions of 670 °/s, which translates into 3 % error. Moreover, our approach does not need a (panoramic) map of the scene to estimate the rotational motion, as opposed to approaches [9, 38]. It also does not need to estimate optical flow prior to fitting a 3D motion, as in [41]. In a nutshell, our approach acts like a visual event-based gyroscope.

### 3.3. Motion Estimation in Planar Scenes

In this section, we show how our framework can be applied to the problem of motion estimation under the assumption of a planar scene (i.e., planar homography estimation), which allows to extract the ego-motion parameters of the camera (rotation and translation) as well as the parameters of the plane containing the scene structure.

In this scenario, image points transform according to $\bar{\mathbf{x}}(t) \propto \mathtt{H}(t)\bar{\mathbf{x}}(0)$, where $\bar{\mathbf{x}} \propto (\mathbf{x}^\top, 1)^\top$ are homogeneous coordinates and $\mathtt{H}(t)$ is a $3 \times 3$ homography matrix. For simplicity, we use $t = 0$ as reference time, and so $\mathtt{H}(0) = \mathtt{Id}$ is the identity. The point trajectories described by $\mathbf{x}(t)$ have the same number of DOFs as $\mathtt{H}(t)$, which, for short time intervals in which we consider $\mathtt{H}$ to be constant, is 8-DOF. We aggregate events along the point trajectories $\mathbf{x}(t)$ defined by candidate homographies $\mathtt{H}(t)$, and maximize the contrast of the resulting image of warped events to recover the homography that best explains the event data.

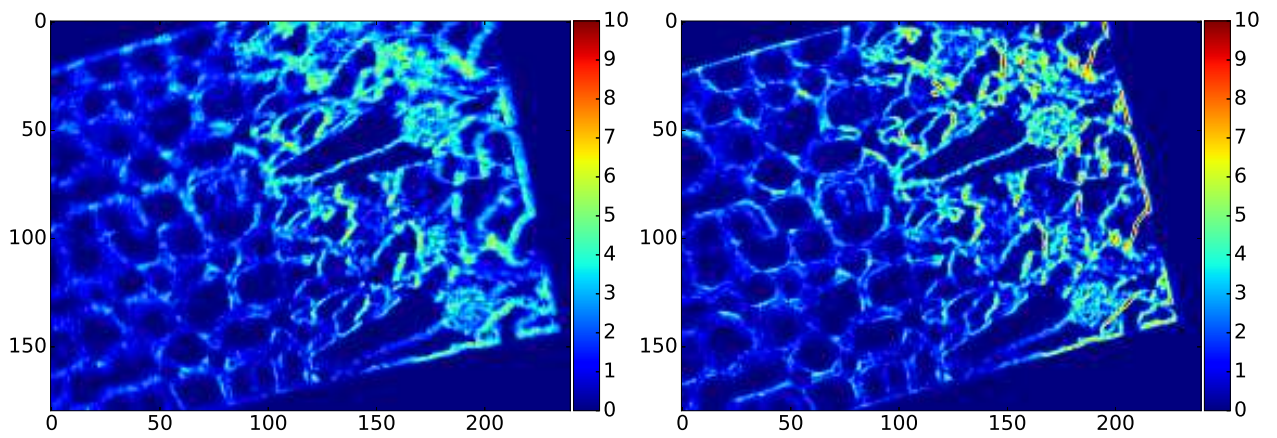

(a) Input events $\mathcal{E}$. (b) Events $\mathcal{E}'$ maximizing contrast.

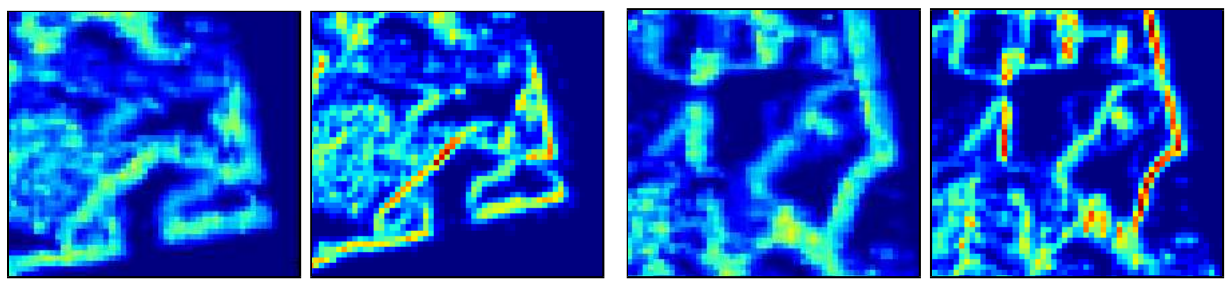

(c) Zoom-in on (a) and (b) for comparison of warped images.

Figure 10: *Motion Estimation in Planar Scenes.* (a) Input events, with each pixel counting the number of events triggered; (b) Warped events using the planar homography parameters $\boldsymbol{\theta} = \{\boldsymbol{\omega}, \mathbf{v}/d, \mathbf{n}\}$ that maximize image contrast: $\boldsymbol{\omega} = (0.086, 0.679, 0.439)^\top$, $\mathbf{v}/d = (0.613, -0.1, 0.333)^\top$, $\mathbf{n} = (0.07, 0.075, -0.995)^\top$.

In case of a homography induced by a plane with homogeneous coordinates $\boldsymbol{\pi} = (\mathbf{n}^\top, d)^\top$, we have [35] $\mathtt{H}(t) \propto \mathtt{R}(t) - \frac{1}{d}\mathbf{t}(t)\mathbf{n}^\top$, where $\mathtt{P}(0) \propto (\mathtt{Id}|\mathbf{0})$ is the projection matrix of the reference view and $\mathtt{P}(t) \propto (\mathtt{R}(t)|\mathbf{t}(t))$ is the projection matrix of the event camera at time $t$. For $t \in [0, \Delta t]$ in a short time interval, we may compute $\mathtt{P}(t)$ from the linear and angular velocities of the event camera, $\mathbf{v}$ and $\boldsymbol{\omega}$, respectively, by assuming that they are constant within the interval: $\mathtt{R}(t) = \exp(\widehat{\boldsymbol{\omega}} t)$ and $\mathbf{t}(t) = \mathbf{v}t$. Hence, we may parametrize $\mathtt{H}(t) \equiv \mathtt{H}(t; \boldsymbol{\theta})$ by $\boldsymbol{\theta} = (\boldsymbol{\omega}^\top, \mathbf{v}^\top/d, \phi, \psi)^\top \in \mathbb{R}^8$, where the 2-DOFs $(\phi, \psi)$ parametrize the unit vector of the plane $\mathbf{n}$ (e.g., latitude-longitude parameters). The parameters $\mathbf{v}/d$ account for the well-known fact that, without additional information, there is a scale ambiguity: the decomposition of a planar homography [35, 39] only provides the direction of the translation, but not its magnitude.

Hence, we consider the events $\mathcal{E}$ in a short time interval, $[0, \Delta t]$, and map them onto the image plane of the reference view using the warp specified in calibrated coordinates by

$$\mathbf{W}(\mathbf{x}_k, t_k; \boldsymbol{\theta}) \propto \mathtt{H}^{-1}(t_k; \boldsymbol{\theta})\, \bar{\mathbf{x}}_k. \qquad (6)$$

Then, the image of warped events is built as usual (2), and its variance (3), i.e., contrast, is computed to assess the quality of the parameters $\boldsymbol{\theta}$ on event alignment.

Fig. 10 shows our method in action. The scene consists of a freely moving event camera viewing a rock poster [31]. In Fig. 10a, a set $\mathcal{E}$ of $N_e = 50\,000$ events is displayed in an event image, with each pixel counting the number of events triggered within it, i.e., as if the identity warp ($\mathbf{x}'_k = \mathbf{x}_k$) was used in (2). Fig. 10b shows the result of contrast maximization: for the optimal parameters $\boldsymbol{\theta}$, the warped events are better aligned with each other, resulting in an image (2) with higher contrast than that in Fig. 10a. Observe that event alignment by contrast maximization produces a motion-corrected image, which is specially noticeable at texture edges: in Fig. 10a (no motion correction) edges are blurred, whereas in Fig. 10b edges are sharp.

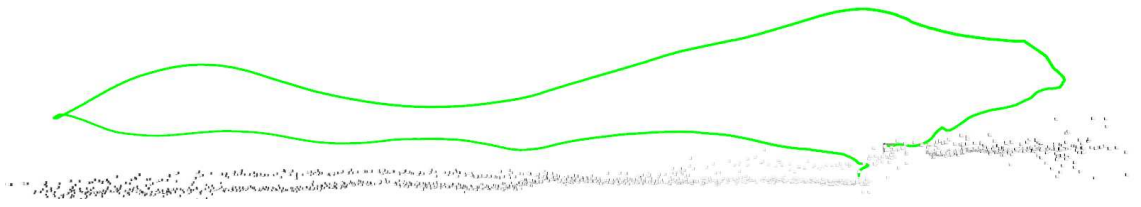

(a) Without motion correction.

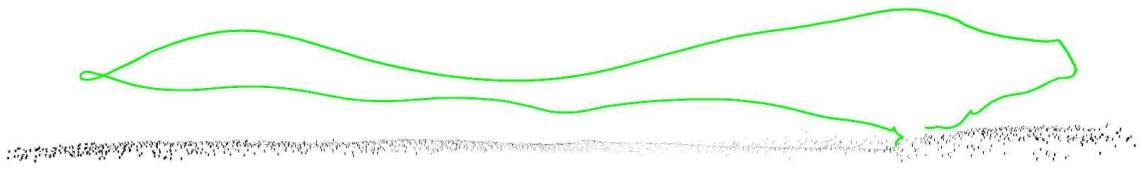

(b) With motion correction.

Figure 11: *Motion Estimation in Planar Scenes.* Scene structure (black dots) and camera motion (green trajectory) obtained by a visual-inertial algorithm [27], with and without motion-corrected event images.

Fig. 11 shows another example of our framework. In this sequence, an event camera is hand-held, looking downwards while a person is walking outdoors over a brick-patterned ground. Event images are used in a visual-inertial algorithm [27] that recovers the trajectory of the event camera and a sparse 3D point map of the scene. The motion-corrected images resulting from homography estimation (cf. Fig. 10b) produce better results, which can be seen by the more flat point cloud representing the floor in the scene.

# 4. Conclusion

In this work, we have focused on showing the capabilities of our framework to tackle several important vision problems for event cameras (3D reconstruction, optical flow and motion estimation), which we believe is its most remarkable characteristic. We showed that there is a simple, principled way to process events in all these problems in the absence of additional appearance information about the scene: maximizing event alignment along point trajectories on the image plane.

Event cameras have multiple desirable properties: low latency, very high dynamic range and low power consumption. We believe this work is a significant step forward into leveraging the advantages of these novel sensors in real-world scenarios, overcoming the limitations of conventional imaging sensors.

# Acknowledgement

This work was supported by the DARPA FLA program, the Swiss National Center of Competence Research Robotics, through the Swiss National Science Foundation, and by the SNSF-ERC starting grant. We thank Alex Z. Zhu and Prof. Kostas Daniilidis for the dataset used in Fig. 11.

## References

[1] Patrick Lichtsteiner, Christoph Posch, and Tobi Delbruck. A 128x128 120dB 30mW asynchronous vision sensor that responds to relative intensity change. In *IEEE Intl. Solid-State Circuits Conf. (ISSCC)*, pages 2060–2069, February 2006. 1

[2] Elias Mueggler, Basil Huber, and Davide Scaramuzza. Event-based, 6-DOF pose tracking for high-speed maneuvers. In *IEEE/RSJ Int. Conf. Intell. Robot. Syst. (IROS)*, pages 2761–2768, 2014. 1, 5

[3] Ryad Benosman, Charles Clercq, Xavier Lagorce, Sio-Hoi Ieng, and Chiara Bartolozzi. Event-based visual flow. *IEEE Trans. Neural Netw. Learn. Syst.*, 25(2):407–417, 2014. 1

[4] Elias Mueggler, Christian Forster, Nathan Baumli, Guillermo Gallego, and Davide Scaramuzza. Lifetime estimation of events from dynamic vision sensors. In *IEEE Int. Conf. Robot. Autom. (ICRA)*, pages 4874–4881, 2015. 1

[5] David Weikersdorfer and Jörg Conradt. Event-based particle filtering for robot self-localization. In *IEEE Int. Conf. Robot. Biomimetics (ROBIO)*, pages 866–870, 2012. 1

[6] Zhenjiang Ni, Aude Bolopion, Joel Agnus, Ryad Benosman, and Stephane Regnier. Asynchronous event-based visual shape tracking for stable haptic feedback in microrobotics. *IEEE Trans. Robot.*, 28:1081–1089, 2012. 1

[7] David Weikersdorfer, Raoul Hoffmann, and Jörg Conradt. Simultaneous localization and mapping for event-based vision systems. In *Int. Conf. Comput. Vis. Syst. (ICVS)*, pages 133–142, 2013. 1

[8] Andrea Censi and Davide Scaramuzza. Low-latency event-based visual odometry. In *IEEE Int. Conf. Robot. Autom. (ICRA)*, pages 703–710, 2014. 1

[9] Hanme Kim, Ankur Handa, Ryad Benosman, Sio-Hoi Ieng, and Andrew J. Davison. Simultaneous mosaicing and tracking with an event camera. In *British Machine Vis. Conf. (BMVC)*, 2014. 1, 6, 7

[10] Guillermo Gallego, Christian Forster, Elias Mueggler, and Davide Scaramuzza. Event-based camera pose tracking using a generative event model. arXiv:1510.01972, 2015. 1

[11] Guillermo Gallego, Jon E. A. Lund, Elias Mueggler, Henri Rebecq, Tobi Delbruck, and Davide Scaramuzza. Event-based, 6-DOF camera tracking from photometric depth maps. *IEEE Trans. Pattern Anal. Machine Intell.*, 2017. 1

[12] David Tedaldi, Guillermo Gallego, Elias Mueggler, and Davide Scaramuzza. Feature detection and tracking with the dynamic and active-pixel vision sensor (DAVIS). In *Int. Conf. Event-Based Control, Comm. Signal Proc. (EBCCSP)*, pages 1–7, 2016. 1

[13] Beat Kueng, Elias Mueggler, Guillermo Gallego, and Davide Scaramuzza. Low-latency visual odometry using event-based feature tracks. In *IEEE/RSJ Int. Conf. Intell. Robot. Syst. (IROS)*, pages 16–23, Daejeon, Korea, October 2016. 1

[14] Hanme Kim, Stefan Leutenegger, and Andrew J. Davison. Real-time 3D reconstruction and 6-DoF tracking with an event camera. In *Eur. Conf. Comput. Vis. (ECCV)*, pages 349–364, 2016. 1

[15] Elias Mueggler, Chiara Bartolozzi, and Davide Scaramuzza. Fast event-based corner detection. In *British Machine Vis. Conf. (BMVC)*, 2017. 1

[16] Jürgen Kogler, Christoph Sulzbachner, and Wilfried Kubinger. Bio-inspired stereo vision system with silicon retina imagers. In *Int. Conf. Comput. Vis. Syst. (ICVS)*, pages 174–183, 2009. 1

[17] Jürgen Kogler, Christoph Sulzbachner, Martin Humenberger, and Florian Eibensteiner. Address-event based stereo vision with bio-inspired silicon retina imagers. In *Advances in Theory and Applications of Stereo Vision*, pages 165–188. InTech, 2011. 1

[18] Ana I. Maqueda, Antonio Loquercio, Guillermo Gallego, Narciso García, and Davide Scaramuzza. Event-based vision meets deep learning on steering prediction for self-driving cars. In *IEEE Int. Conf. Comput. Vis. Pattern Recog. (CVPR)*, 2018. 1

[19] Patrick Bardow, Andrew J. Davison, and Stefan Leutenegger. Simultaneous optical flow and intensity estimation from an event camera. In *IEEE Int. Conf. Comput. Vis. Pattern Recog. (CVPR)*, pages 884–892, 2016. 1

[20] Henri Rebecq, Guillermo Gallego, and Davide Scaramuzza. EMVS: Event-based multi-view stereo. In *British Machine Vis. Conf. (BMVC)*, September 2016. 2

[21] Henri Rebecq, Guillermo Gallego, Elias Mueggler, and Davide Scaramuzza. EMVS: Event-based multi-view stereo—3d reconstruction with an event camera in real-time. *Int. J. Comput. Vis.*, pages 1–21, November 2017. 2, 4, 5, 6

[22] Guillermo Gallego and Davide Scaramuzza. Accurate angular velocity estimation with an event camera. *IEEE Robot. Autom. Lett.*, 2:632–639, 2017. 2, 6, 7

[23] Alex Zihao Zhu, Nikolay Atanasov, and Kostas Daniilidis. Event-based feature tracking with probabilistic data association. In *IEEE Int. Conf. Robot. Autom. (ICRA)*, pages 4465–4470, 2017. 2

[24] Elias Mueggler, Guillermo Gallego, and Davide Scaramuzza. Continuous-time trajectory estimation for event-based vision sensors. In *Robotics: Science and Systems (RSS)*, 2015. 2

[25] Elias Mueggler, Guillermo Gallego, Henri Rebecq, and Davide Scaramuzza. Continuous-time visual-inertial odometry with event cameras. *IEEE Trans. Robot.*, 2017. Under review. 2

[26] Alex Zihao Zhu, Nikolay Atanasov, and Kostas Daniilidis. Event-based visual inertial odometry. In *IEEE Int. Conf. Comput. Vis. Pattern Recog. (CVPR)*, pages 5816–5824, 2017. 2

[27] Henri Rebecq, Timo Horstschaefer, and Davide Scaramuzza. Real-time visual-inertial odometry for event cameras using keyframe-based nonlinear optimization. In *British Machine Vis. Conf. (BMVC)*, 2017. 2, 3, 8

[28] Bruce D. Lucas and Takeo Kanade. An iterative image registration technique with an application to stereo vision. In *Int. Joint Conf. Artificial Intell. (IJCAI)*, pages 674–679, 1981. 3

[29] Antoni Rosinol Vidal, Henri Rebecq, Timo Horstschaefer, and Davide Scaramuzza. Ultimate SLAM? combining events, images, and IMU for robust visual SLAM in HDR and high speed scenarios. *IEEE Robot. Autom. Lett.*, 3(2):994–1001, April 2018. 3

[30] Anthony Yezzi, Andy Tsai, and Alan Willsky. A statistical approach to snakes for bimodal and trimodal imagery. In *Int. Conf. Comput. Vis. (ICCV)*, volume 2, pages 898–903, 1999.

4

[31] Elias Mueggler, Henri Rebecq, Guillermo Gallego, Tobi Delbruck, and Davide Scaramuzza. The event-camera dataset and simulator: Event-based data for pose estimation, visual odometry, and SLAM. *Int. J. Robot. Research*, 36:142–149, 2017. 4, 5, 6, 7, 8

[32] Henri Rebecq, Timo Horstschäfer, Guillermo Gallego, and Davide Scaramuzza. EVO: A geometric approach to event-based 6-DOF parallel tracking and mapping in real-time. *IEEE Robot. Autom. Lett.*, 2:593–600, 2017. 4

[33] Christian Brandli, Raphael Berner, Minhao Yang, Shih-Chii Liu, and Tobi Delbruck. A 240x180 130dB 3us latency global shutter spatiotemporal vision sensor. *IEEE J. Solid-State Circuits*, 49(10):2333–2341, 2014. 5

[34] Zhengyou Zhang. A flexible new technique for camera calibration. *IEEE Trans. Pattern Anal. Machine Intell.*, 22(11):1330–1334, November 2000. 5

[35] Richard Hartley and Andrew Zisserman. *Multiple View Geometry in Computer Vision*. Cambridge University Press, 2003. Second Edition. 5, 7, 8

[36] Rafael C. Gonzalez and Richard Eugene Woods. *Digital Image Processing*. Pearson Education, 2009. 6

[37] Matthew Cook, Luca Gugelmann, Florian Jug, Christoph Krautz, and Angelika Steger. Interacting maps for fast visual interpretation. In *Int. Joint Conf. Neural Netw. (IJCNN)*, pages 770–776, 2011. 6

[38] Christian Reinbacher, Gottfried Munda, and Thomas Pock. Real-time panoramic tracking for event cameras. In *IEEE Int. Conf. Comput. Photography (ICCP)*, pages 1–9, 2017. 6, 7

[39] Yi Ma, Stefano Soatto, Jana Košecká, and Shankar Sastry. *An Invitation to 3-D Vision: From Images to Geometric Models*. Springer, 2004. 7, 8

[40] William W Hager and Hongchao Zhang. A survey of nonlinear conjugate gradient methods. *Pacific J. of Optimization*, 2(1):35–58, 2006. 7

[41] Jörg Conradt. On-board real-time optic-flow for miniature event-based vision sensors. In *IEEE Int. Conf. Robot. Biomimetics (ROBIO)*, pages 1858–1863, 2015. 7

# A Unifying Contrast Maximization Framework for Event Cameras, with Applications to Motion, Depth, and Optical Flow Estimation – Supplementary Material –

Guillermo Gallego, Henri Rebecq, Davide Scaramuzza
Dept. of Informatics and Neuroinformatics, University of Zurich and ETH Zurich
http://rpg.ifi.uzh.ch

## 5. Multimedia Material

A video showing the application of our framework to solve several computer vision problems with event cameras is available at: https://youtu.be/KFMZFhi-9Aw.

## 6. Optical Flow Estimation

Our framework seeks for the point trajectories on the image plane that best fit the event data and it is able to take into account all the information contained in the events: space-time coordinates and polarity (i.e., sign) of the brightness changes. More specifically, event polarity is incorporated in the framework during the creation of the image patches of warped events (equation (2) in the paper).

Figure 12 compares the elements of our framework (image patches of warped events $H$ and objective function $f$) for a set of events whose optical flow we want to estimate, in two scenarios:

1. Using polarity ($b_k = p_k$), i.e.,

$$H(\mathbf{x};\theta) = \sum_{k=1}^{N_e} p_k\, \delta(\mathbf{x} - \mathbf{x}'_k(\theta)). \tag{7}$$

2. Not using polarity ($b_k = 1$), i.e.,

$$H(\mathbf{x};\theta) = \sum_{k=1}^{N_e} \delta(\mathbf{x} - \mathbf{x}'_k(\theta)). \tag{8}$$

For illustration purposes, the intensity frame in Fig. 12a shows the patch corresponding to the considered events (yellow rectangle). However, such an intensity frame is not used in our framework.

Without using polarity, Figs. 12b and 12c show the contrast function $f(\theta)$ and images of warped events (8) for three candidate point trajectories, specified by the three optical flow vectors $\theta_i, i = 0,1,2$, that are displayed in Fig. 12a. Conversely, Figs. 12d and 12e show the corresponding elements if event polarity is used. The image patches of warped events are color coded from blue (low) to red (high). If polarity is not used (Fig. 12b), blue means absence of events (small values of (8)), whereas red indicates large accumulation of events (large values of (8)). If polarity is used (Fig. 12e), green means absence of events, whereas red and blue indicate large accumulation of positive and negative events, respectively, according to (7).

As it can be observed by comparing Figs. 12c and 12d, both objective functions provide approximately the same optimal velocity (peak of the objective function) $\theta \equiv \mathbf{v} \approx (-40, 0)$ pixel/s. However, the basin of attraction of the optimal value is slightly narrower and more pronounced if polarity is used than if it is not taken into account, as can be noted since Figs. 12c and 12d are displayed using the same color range. This can be explained by comparing the image patches of warped events in Figs. 12b and 12e. In case of thin edge structures like the ones in the considered patch, if events are warped so that nearby edges overlap, and therefore their opposite event polarities cancel, then the contrast function $f(\theta)$ greatly decreases (thus reducing the width of the contrast peak, i.e., its basin of attraction). Conversely, if event polarity is not used, the alignment of nearby edges does not produce cancellation, and therefore the contrast decreases more slowly, due to the warped edges becoming further apart.

## 7. Depth Estimation

To illustrate how depth estimation improves as more events are processed, we carried out an experiment with the `slider_depth` sequence from the dataset [1]. We reconstructed the scene with the same steps as those used for Fig. 7 in the paper, but varying the number of events processed, $N_e$, between 20000 and 1 million. The results are displayed in Fig. 13. As it can be seen, as more events are processed (corresponding to a larger camera baseline), the reconstructed point cloud becomes more accurate and less noisy. This effect is also visible in the semi-dense depth maps overlaid on the grayscale frame of the reference view.

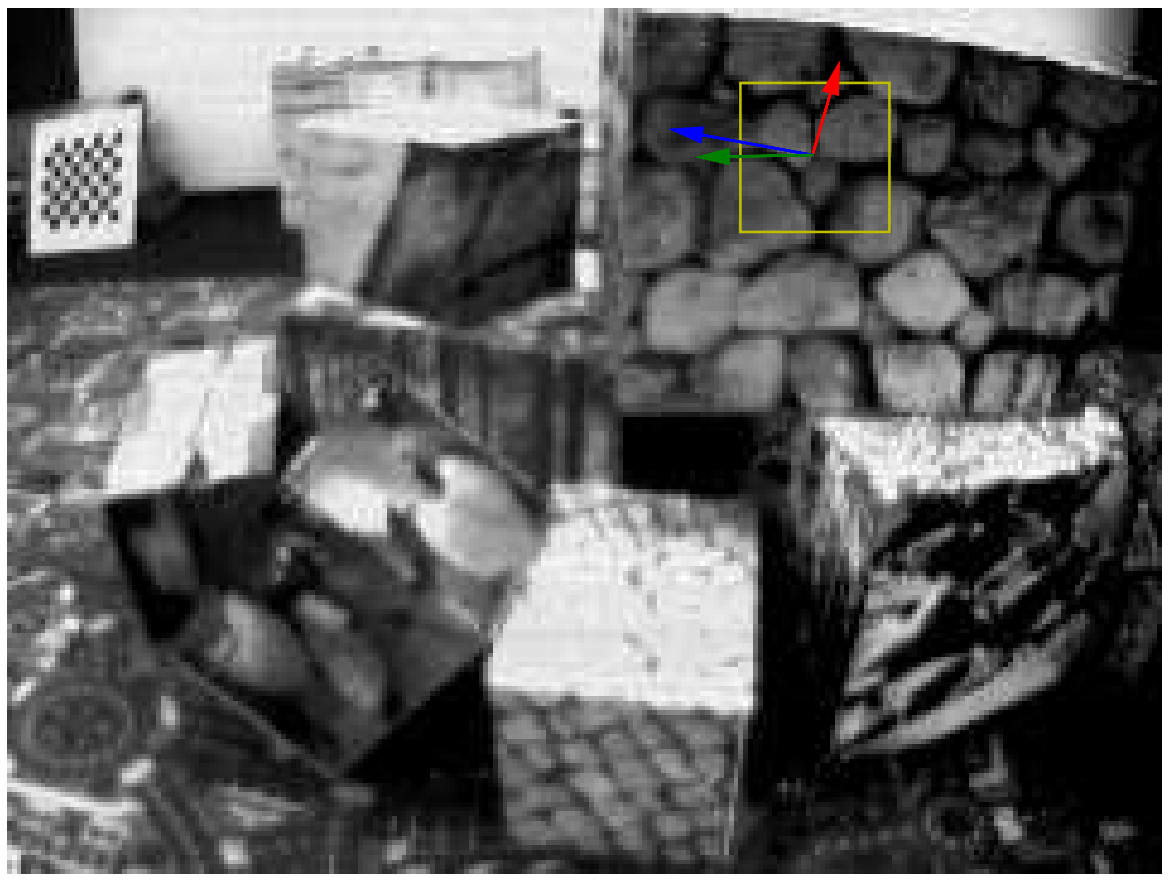

(a) Intensity frame showing the patch corresponding to the considered events (yellow rectangle) and three candidate velocities (optical flow): $\theta_0$ (red arrow), $\theta_1$ (blue arrow) and $\theta_2$ (green arrow).

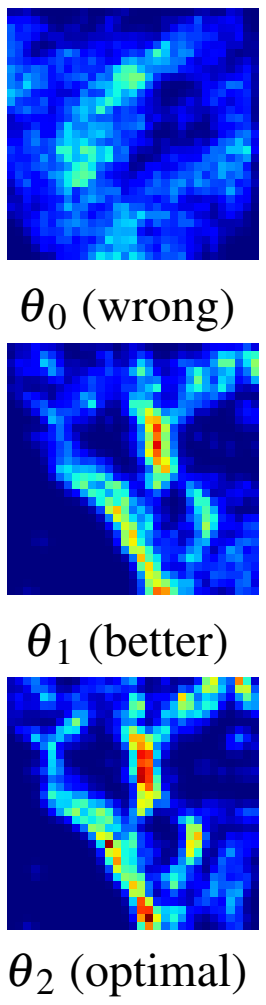

(b) Warped events $H(\mathbf{x};\theta)$, for $\{\theta_i\}_{i=0}^{2}$ in Fig. 12c.

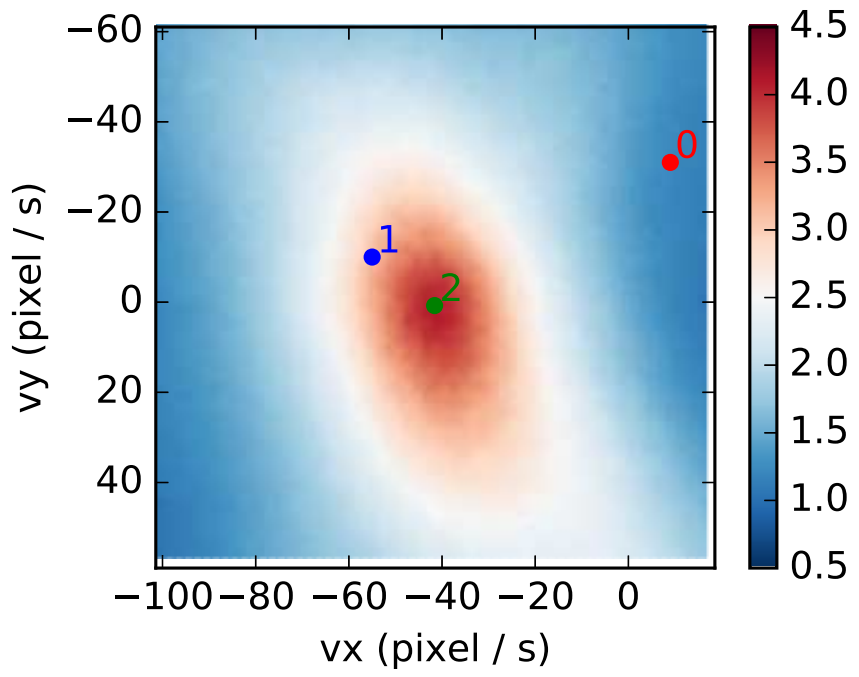

(c) Contrast function $f(\theta)$ for $H$ in (8). For illustration, we show in Fig. 12b the images of warped events (8) corresponding to the three candidate velocities in Fig. 12a.

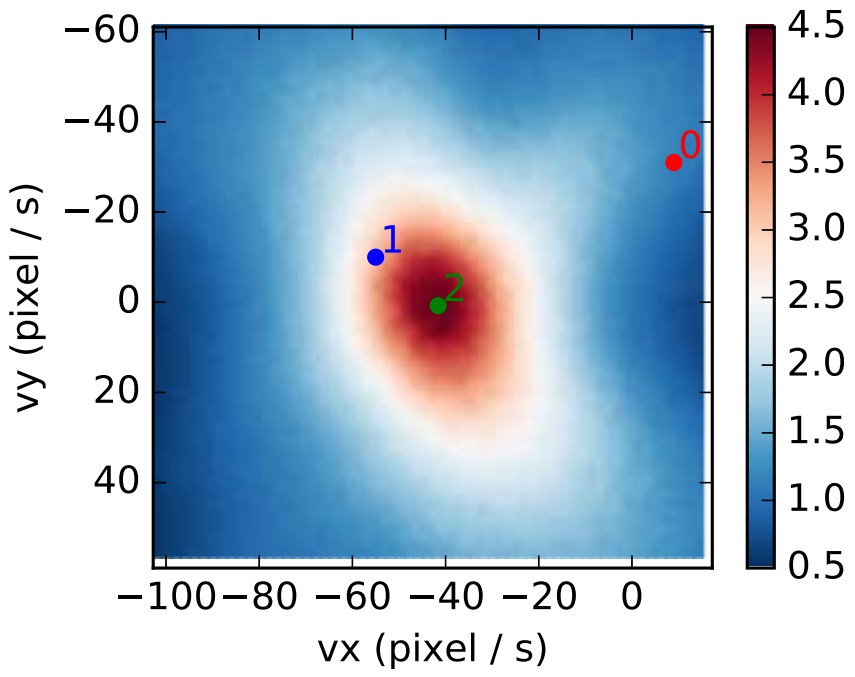

(d) Contrast function $f(\theta)$ for $H$ in (7). For illustration, we show in Fig. 12e the images of warped events (7) corresponding to the three candidate velocities in Fig. 12a.

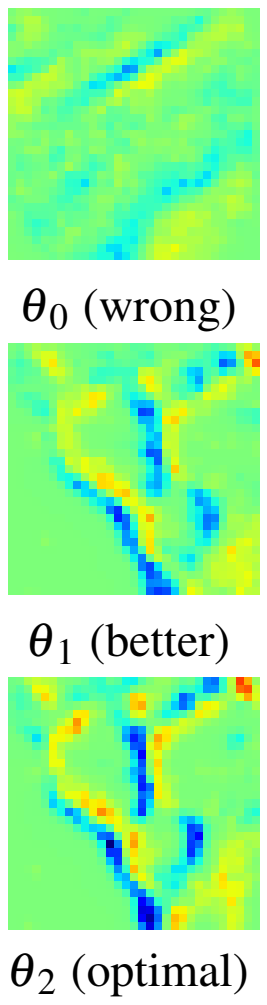

(e) Warped events $H(\mathbf{x};\theta)$, for $\{\theta_i\}_{i=0}^{2}$ in Fig. 12d.

Figure 12: *Optical Flow* (patch-based) estimation. Comparison of objective functions and images of warped events using event polarity (Figs. 12e and 12d) or not using it (Figs. 12b and 12c). In either case, the optical flow is estimated by finding the maximizer of the contrast $f(\theta)$.

# 8. Rotational Motion Estimation

A comparison between the types of warped event images $H$ obtained depending on whether they store the event count ($b_k = 1$) or the balance of polarities ($b_k = p_k$) is shown in Fig. 14.

In the top row of Fig. 14, polarity is not used. For visualization purposes, event images in this row (Figs. 14a and 14b) are displayed in negative form (bright means lack of events and dark means abundance of events)[1]. As it is observed, per-pixel event accumulation (Fig. 14a) produces a motion-blurred image since events are triggered by moving edges. Contrarily, the image of warped events using the estimated motion parameters (Fig. 14b) presents a higher contrast and sharpness than the image in Fig. 14a, which indicates a better *alignment of the events* along the candidate

[1]Figs. 14a and 14b are the same type of images as in Fig. 12b, but with a different color scheme: from white to black instead of from blue to read.

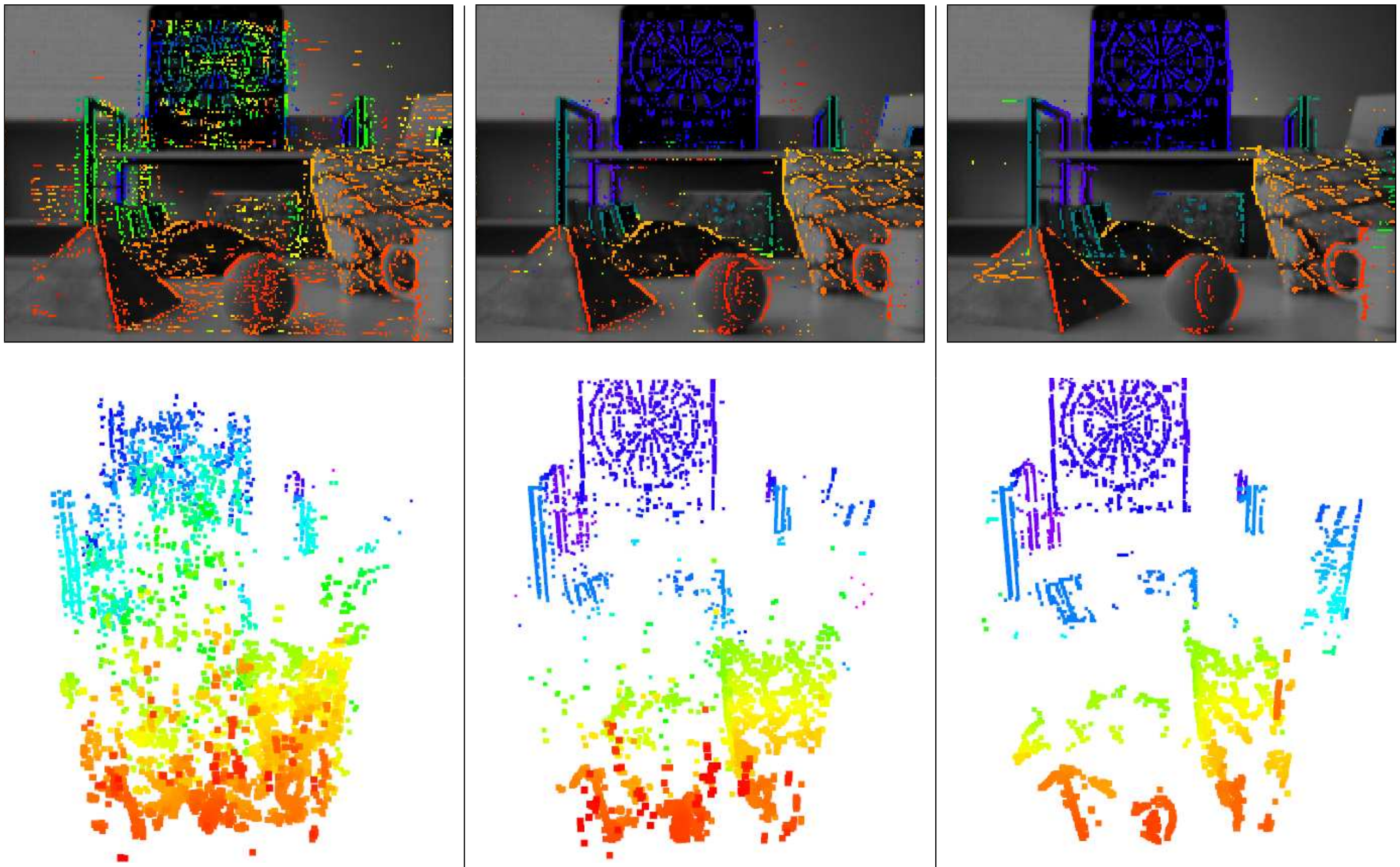

(a) $N_e = 20\,000$ events
Baseline: 1.9 cm
Time span: 0.06 s.

(b) $N_e = 100\,000$ events
Baseline: 9.5 cm
Time span: 0.33 s.

(c) $N_e = 1\,000\,000$ events
Baseline: 85 cm
Time span: 2.93 s.

Figure 13: *Depth Estimation* for different subsets of events, with increasing time span and baseline. Top row: depth map overlaid on grayscale frame. Bottom row: 3D reconstruction (point cloud). Depth is color-coded, from red (close) to blue (far), in the range of 0.45 m to 2.4 m.

point trajectories on the image plane. The effect of having a higher contrast can also be noticed by comparing the distribution of values (i.e., histogram) of the images of warped events, as shown in Fig. 14c. The image with larger contrast (Fig. 14b) has a larger *range of values* than the image with lower contrast (i.e., darker pixels and larger amount of dark pixels in Fig. 14b with respect to Fig. 14a), and, since the range of values is non-negative and with a peak at zero, this means that the mass distribution of values shifts toward larger (positive) numbers as the contrast increases (i.e., the red curve in Fig. 14c becomes the blue curve as contrast increases).

The previous observations are also applicable to the second row of Fig. 14, where *event polarity* is used (cf. Figs. 14d and 14e). The average gray level corresponds to pixels where no events were generated; dark regions correspond to negative events, and bright regions correspond to positive events. Indeed, the image of warped events $H$ obtained with the optimal parameters (Fig. 14e) has a larger contrast than the one with per-pixel polarity accumulation (Fig. 14d). The larger contrast of Fig. 14e over Fig. 14d is evidenced by the larger range of values and larger amount of brighter and darker pixels, as reported in the comparison of the distributions (Fig. 14f) of pixel values in both images.

We quantified the effect of using or not using the event polarity for rotational motion estimation on sequences from the dataset [1]. Each sequence has a 1 minute length and contains about 100-200 million events. Ground truth camera motion is provided by a sub-millimeter motion capture system. Each rotational motion sequence starts with rotations around each camera axis, and then is followed by rotations in all 3-DOFs. In addition, the speed of the motion increases as the sequence progresses. Fig. 15 shows the comparison of the results of our framework, not using event polarity ($b_k = 1$), against ground truth on the `poster_rotation` sequence. The curves corresponding to the 3-DOFs of the event camera on the entire sequence are shown in Fig. 15a. This plot shows the increasing speed of the motion, with excitations close to $\pm 1000\,^\circ/\mathrm{s}$. Figures 15b and 15c are zoomed-in versions of Fig. 15a, with rotations dominantly around each axis of the event camera (Fig. 15b) or in arbitrary axes (Fig. 15c), respectively. Our framework provides very accurate results, as highlighted by the very small errors: the lines of our method and those of the ground truth are almost indistinguishable at this scale. These errors are better noticed in the boxplots of Fig 16a, where errors are reported in sub-intervals of 15 s, in accordance with the increasing speed of the motion in the sequence. Fig. 16b reports the boxplot errors in case of using the event polarity ($b_k = p_k$) to build the image of warped events. As it can be observed by comparing both boxplots (Figs. 16a and 16b), using event polarity does not significantly change the results in this scenario. We measured Root Mean Square (RMS) angular velocity errors over the entire sequence of: $25.96\,^\circ/\mathrm{s}$ (without using polarity) and $24.39\,^\circ/\mathrm{s}$ (using polarity). Both are relatively small, compared to the peak velocities close to $1000\,^\circ/\mathrm{s}$, i.e., in the order of 2.5 % error.

## References

[1] Elias Mueggler, Henri Rebecq, Guillermo Gallego, Tobi Delbruck, and Davide Scaramuzza. The event-camera dataset and simulator: Event-based data for pose estimation, visual odometry, and SLAM. *Int. J. Robot. Research*, 36:142–149, 2017. 1, 3, 5, 6

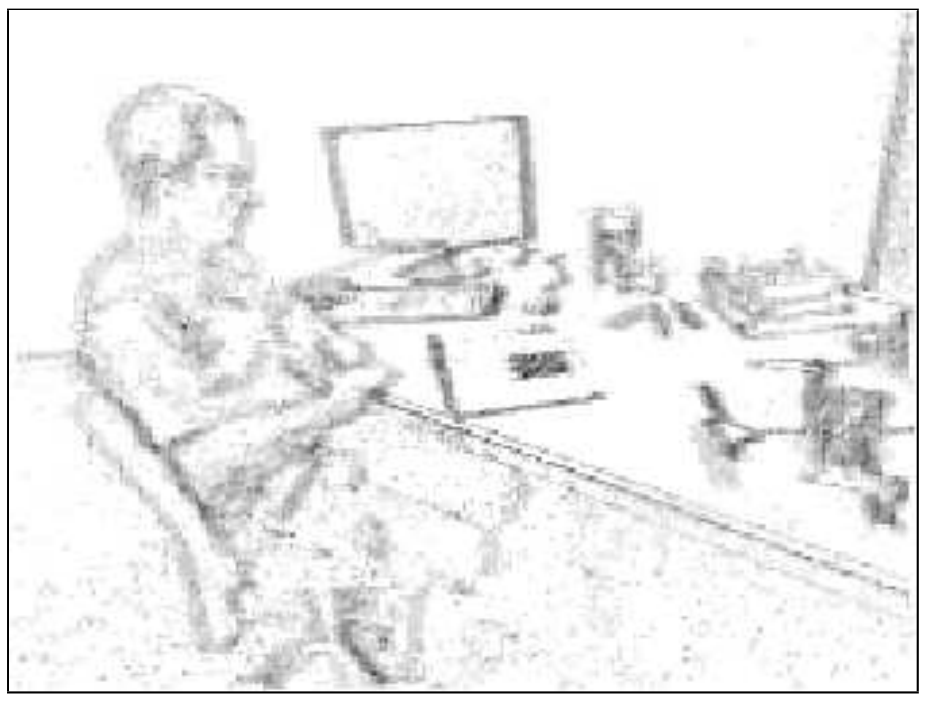

(a) Warped events for angular velocity $\theta = \mathbf{0}$ (i.e., no motion correction). Using $b_k = 1$ in the image of warped events $H$.

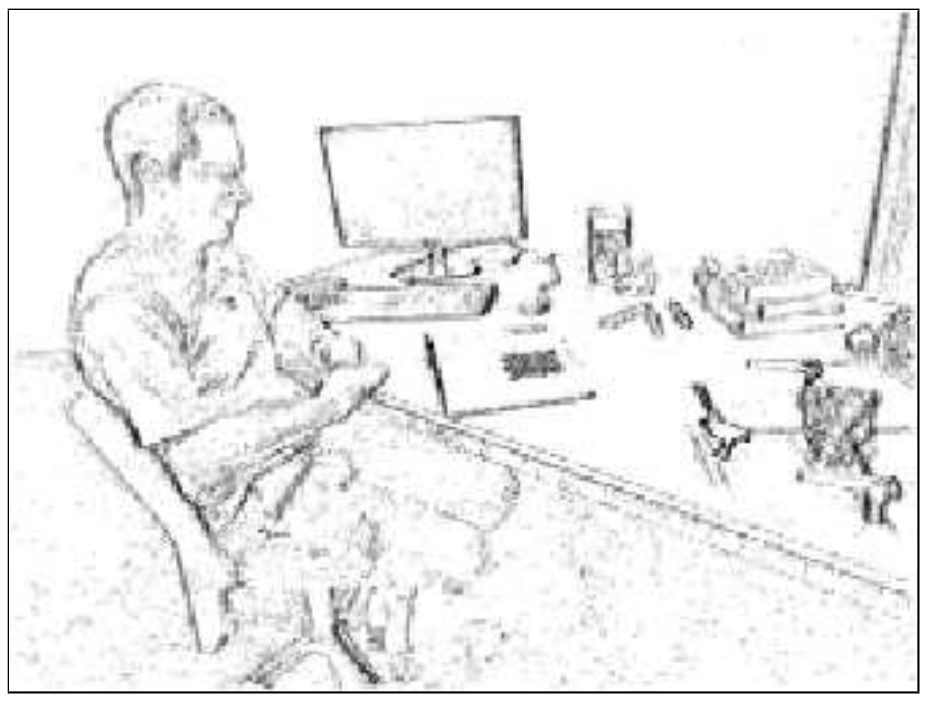

(b) Warped events using the estimated angular velocity $\theta^*$, which produces motion-corrected, sharp edges.

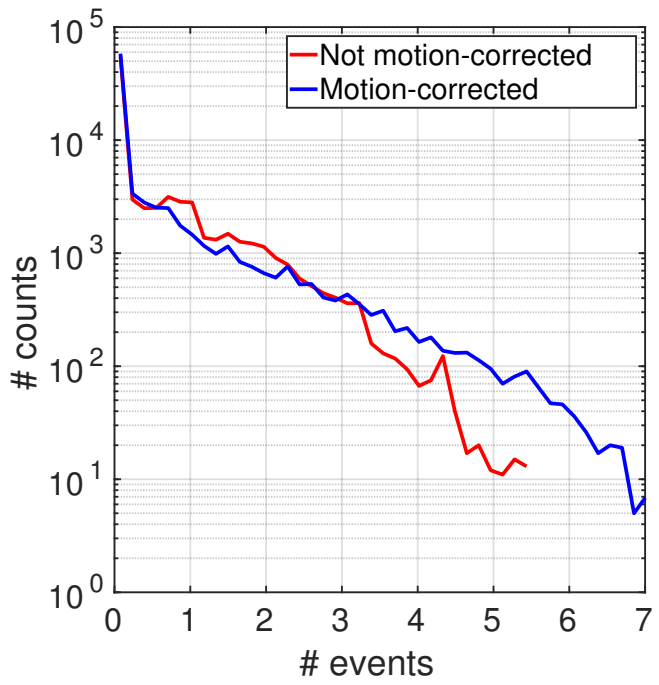

(c) Histograms of the negative of the images in Figs. 14a and 14b. The peak at zero corresponds to the white pixels.

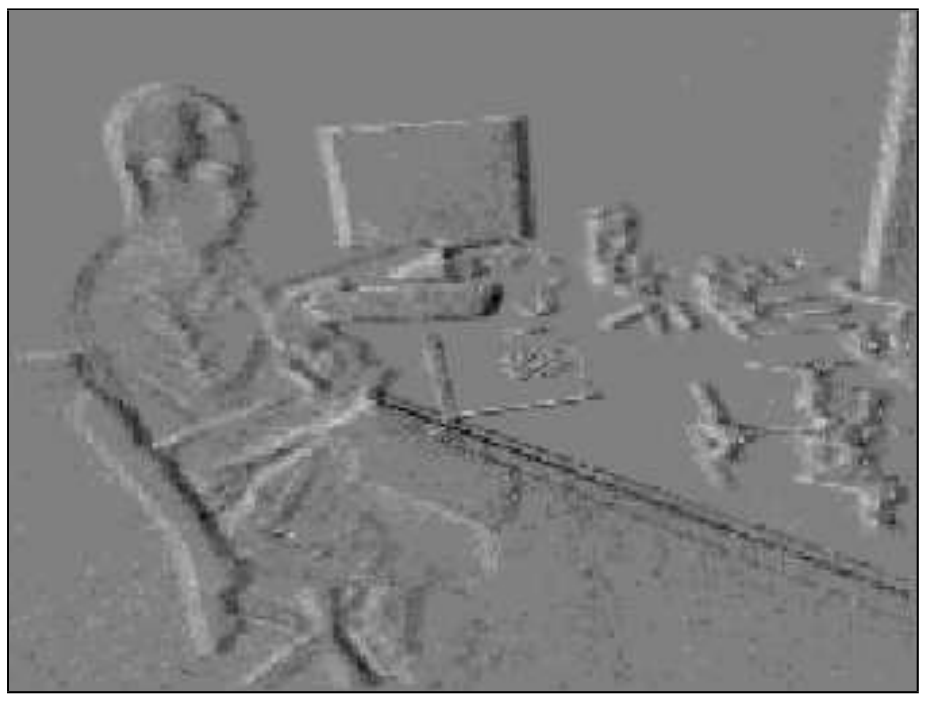

(d) Warped events angular velocity $\theta = \mathbf{0}$ (i.e., no motion correction). Using polarity, $b_k = p_k$, in the image of warped events $H$.

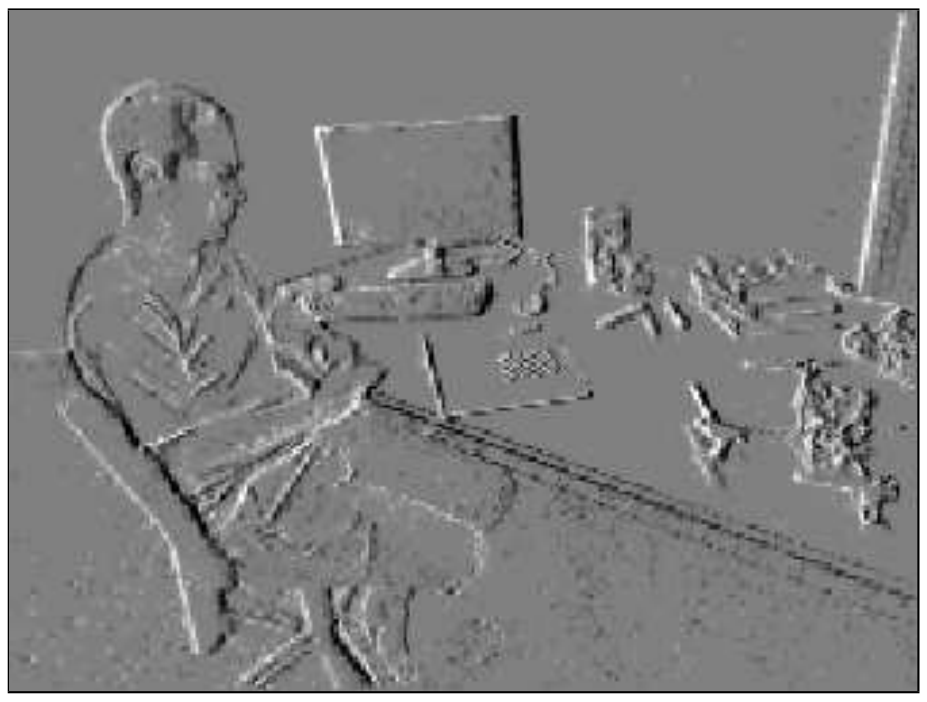

(e) Warped events using the estimated angular velocity $\theta^*$, which produces motion-corrected, sharp edges. Using polarity, $b_k = p_k$, in $H$.

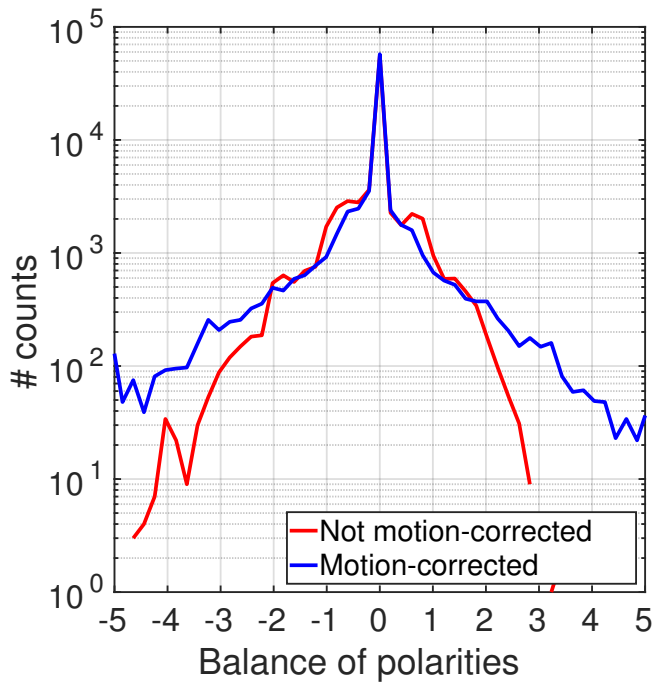

(f) Histograms of Figs. 14d and 14e. The peak at zero corresponds to the gray pixels.

Figure 14: *Rotational Motion Estimation*. Images of warped events, displayed in grayscale to better visualize the motion blur due to event misalignment and the sharpness due to event alignment. Top: Not using polarity ($b_k = 1$ in the image of warped events $H$); bottom: using polarity ($b_k = p_k$ in $H$). Sequence `dynamic_rotation` from the dataset [1].

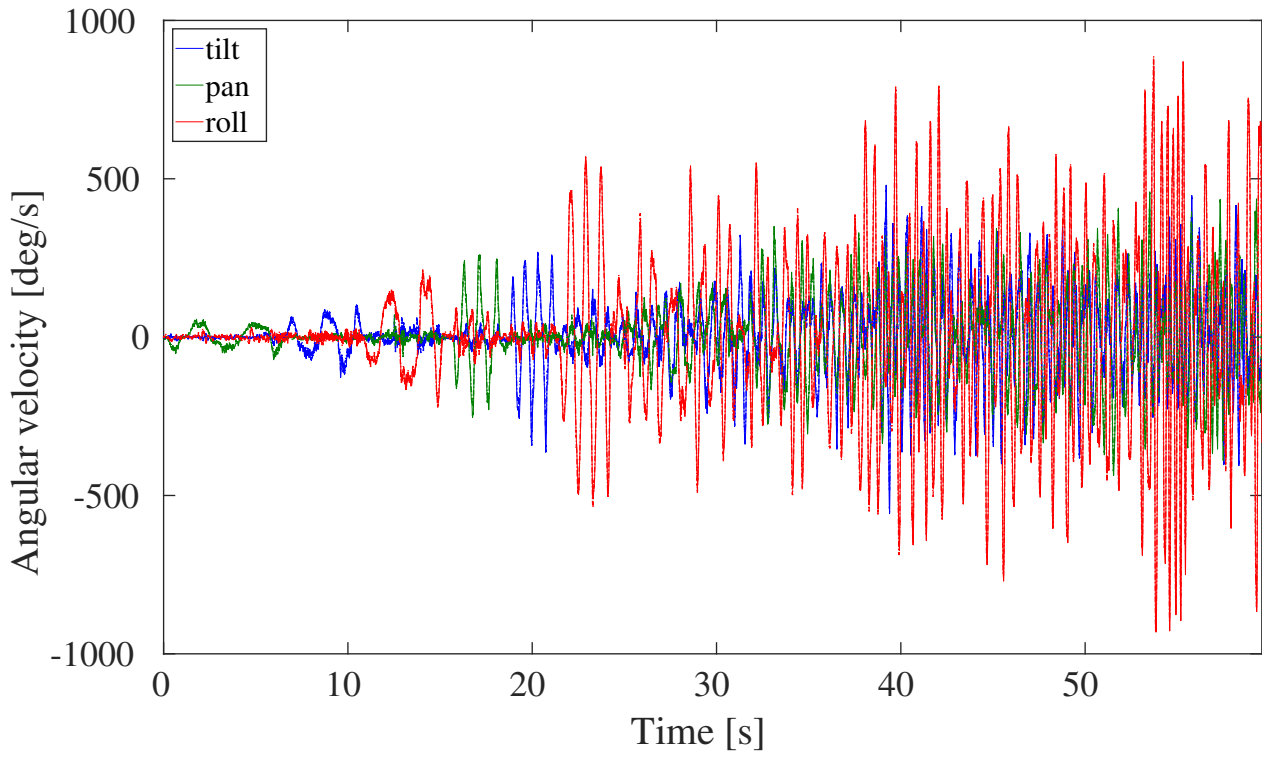

(a) Whole sequence. Rotational motion with increasing velocity, reaching speeds close to $\pm 1000\,^\circ/\mathrm{s}$.

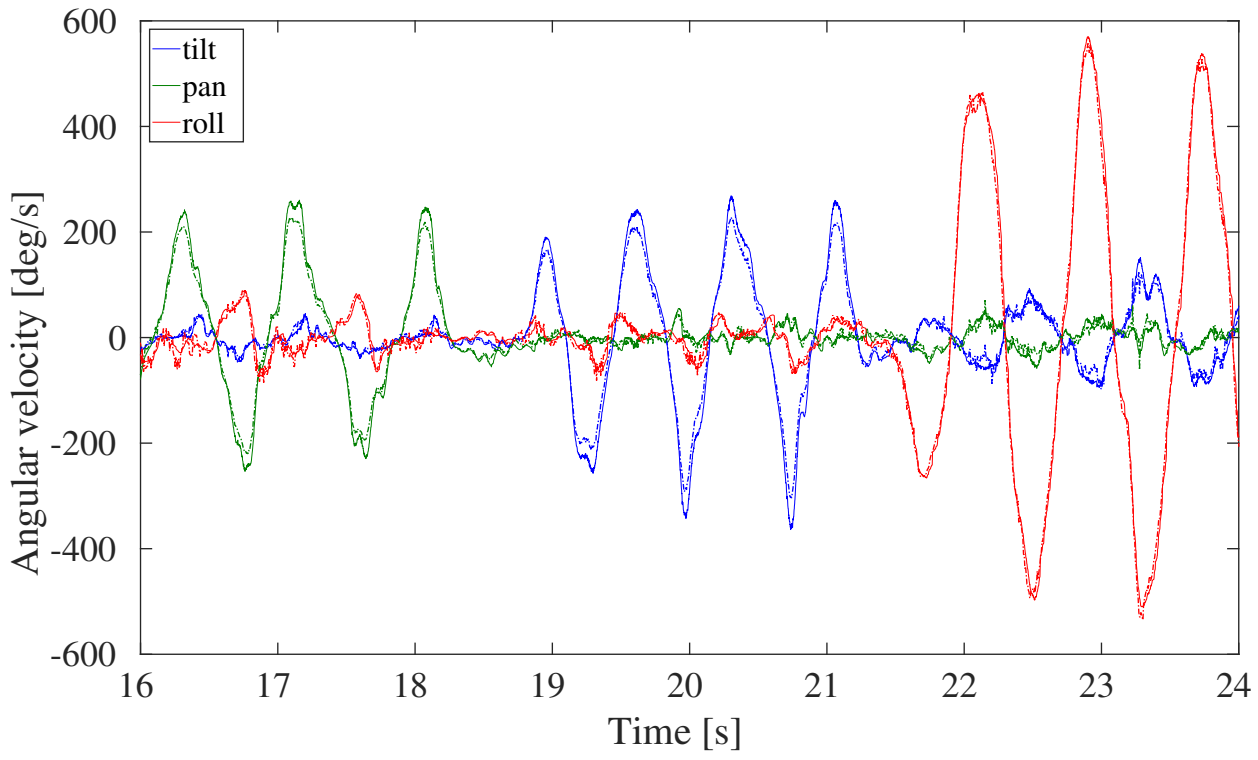

(b) Zoom of Fig. 15a, showing a series of rotations dominantly along one axis: pan (rotation around $Y$ axis), tilt (rotation around $X$ axis) and roll (rotation around $Z$ axis).

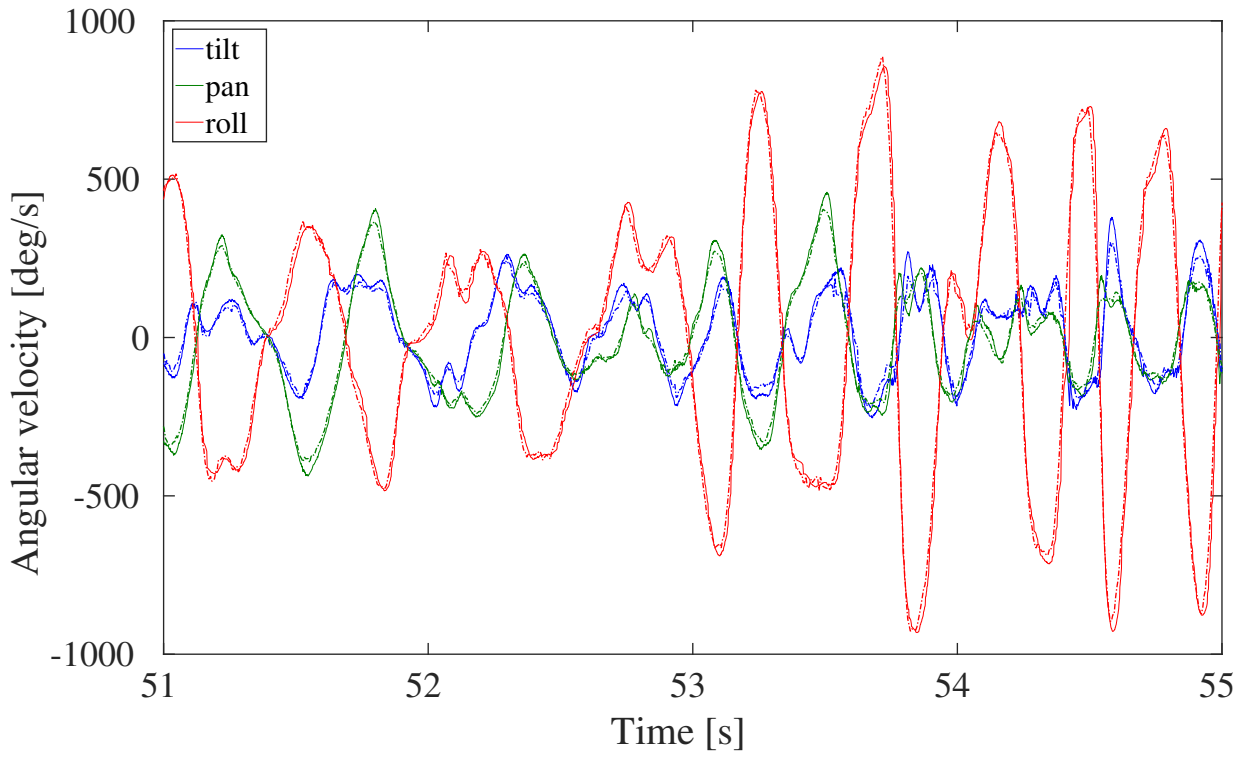

(c) Zoom of Fig. 15a, showing rotations in arbitrary directions, with speed close to $1000\,^\circ/\mathrm{s}$.

Figure 15: *Rotational Motion Estimation*. Comparison of the estimated angular velocity (solid line) using our framework with $b_k = 1$ (i.e., without event polarity) against ground truth (dashed line). Sequence `poster_rotation` in the dataset [1].

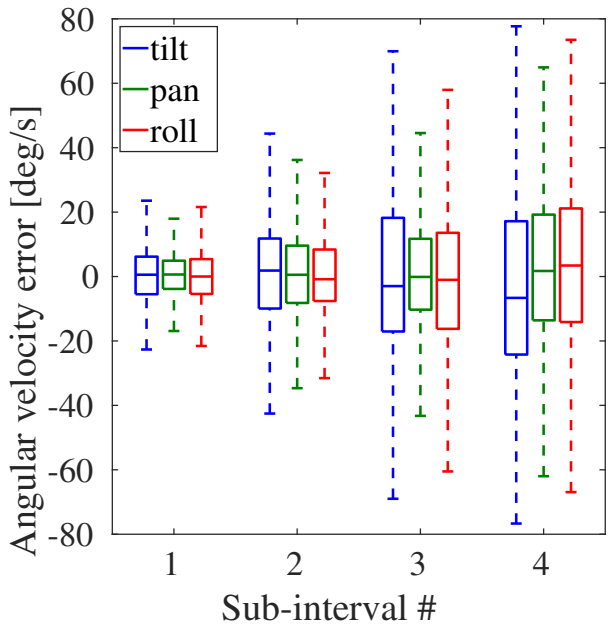

(a) Without polarity

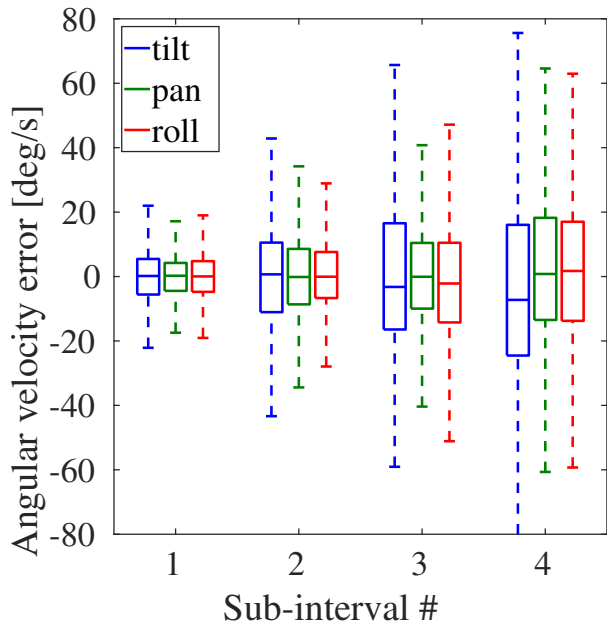

(b) With polarity

Figure 16: *Rotational Motion Estimation*. Angular velocity error (estimated vs. ground truth) for the same sequence, with or wihout taking into account event polarity in the image of warped events $H$. Sequence `poster_rotation` from the dataset [1].